\documentclass[journal]{IEEEtran}

\usepackage{amsmath,amssymb,amsthm}
\usepackage{graphicx}
\usepackage{cite}

\usepackage{subfigure}
\usepackage{float}
\usepackage{bm}

\usepackage{booktabs,multirow}
\usepackage{makecell}
\usepackage{algorithm,algorithmicx,algpseudocode}

\usepackage{verbatim}
\usepackage{tikz}
\usetikzlibrary{spy}
\usepackage{url}

\usepackage{color}

\newcommand{\x}		{\mathbf{x}}	
\newcommand{\y}		{\mathbf{y}}
\newcommand{\A}		{\mathbf{A}}
\newcommand{\z}	 {\mathbf{z}}
\newcommand{\W}	 {\mathbf{W}}
\renewcommand{\P}	 {\mathbf{P}}
\newcommand{\omg}	 {\mathbf{\Omega}}
\newcommand{\X}		{\mathbf{X}}	
\newcommand{\Z}		{\mathbf{Z}}
\newcommand{\0}	     {\mathbf{0}}

\newcommand{\D}	 {\mathbf{D}}
\newcommand{\I}	     {\mathbf{I}}

\newcommand{\LL}	 {\mathbf{L}}
\newcommand{\QQ}	     {\mathbf{Q}}
\newcommand{\RR}	     {\mathbf{R}}
\newcommand{\SSS}	     {\mathbf{\Sigma}}

\newcommand{\s}	 {\mathbf{s}}
\newcommand{\g}	 {\mathbf{g}}
\newcommand{\h}	 {\mathbf{h}}
\newcommand{\ze}	 {\bm{\zeta}}
\newcommand{\kap}	 {\bm{\kappa}}

\newcommand{\R}	     {\mathsf{R}}

\newcommand{\e}		{\mathbf{e}}	
\newcommand{\diag}  {\mathsf{diag}}

\title{PWLS-ULTRA: An Efficient Clustering and Learning-Based Approach for Low-Dose 3D CT Image Reconstruction}
\author{ Xuehang Zheng, Saiprasad Ravishankar, \textit{Member, IEEE}, Yong Long*, \textit{Member, IEEE}, \\ and Jeffrey A. Fessler, \textit{Fellow, IEEE}
	\thanks{ 
		Copyright (c) 2017 IEEE. Personal use of this material is permitted. However, permission to use this material for any other purposes must be obtained from the IEEE by sending a request to pubs-permissions@ieee.org.
		
		This work was supported in part by the SJTU-UM Collaborative Research Program, NSFC (61501292), Shanghai Pujiang Talent Program (15PJ1403900), NIH grant U01 EB018753, ONR grant N00014-15-1-2141, DARPA Young Faculty Award D14AP00086, and ARO MURI grants W911NF-11-1-0391 and 2015-05174-05. \textit{Asterisk indicates the corresponding author.} 
		
		X. Zheng and Y. Long are with the University of Michigan - Shanghai Jiao Tong University Joint Institute, Shanghai Jiao Tong University, Shanghai 200240, China (email: zhxhang@sjtu.edu.cn, yong.long@sjtu.edu.cn).
		
		S. Ravishankar and J. A. Fessler are with the Department of Electrical Engineering and Computer Science, University of Michigan, Ann Arbor, MI, 48109 USA (email: ravisha@umich.edu, fessler@umich.edu).
		
		Digital Object Identifier 10.1109/TMI.2018.2832007.
}}
	
\begin{document}
	\maketitle	
	
\begin{abstract}
	The development of computed tomography (CT) image reconstruction methods that significantly reduce patient radiation exposure while maintaining high image quality is an important area of research in low-dose CT (LDCT) imaging. 
	We propose a new penalized weighted least squares (PWLS) reconstruction method that exploits regularization based on an efficient Union of Learned TRAnsforms (PWLS-ULTRA).
	The union of square transforms is pre-learned from numerous image patches extracted from a dataset of CT images or volumes.
	The proposed PWLS-based cost function is optimized by alternating between a CT image reconstruction step, and a sparse coding and clustering step.
	The CT image reconstruction step is accelerated by a relaxed linearized augmented Lagrangian method with ordered-subsets that reduces the number of forward and back projections.
	Simulations with 2D and 3D axial CT scans of the extended cardiac-torso phantom and 3D helical chest and abdomen scans show that for both normal-dose and low-dose levels, the proposed method significantly improves the quality of reconstructed images compared to PWLS reconstruction with a nonadaptive edge-preserving regularizer (PWLS-EP). 
	PWLS with regularization based on a union of learned transforms leads to better image reconstructions than using a single learned square transform. We also incorporate patch-based weights in PWLS-ULTRA that enhance image quality and help improve image resolution uniformity. The proposed approach achieves comparable or better image quality compared to learned overcomplete synthesis dictionaries, but importantly, is much faster (computationally more efficient). 
\end{abstract}	

\begin{IEEEkeywords}
	Low-dose CT, statistical image reconstruction, sparse representations, sparsifying transform learning, dictionary learning, machine learning.
\end{IEEEkeywords}

\section{Introduction}     

There is a growing interest in techniques for computed tomography (CT) image reconstruction that significantly reduce patient radiation exposure while maintaining high image quality.
Dictionary learning based techniques have been proposed for low-dose CT (LDCT) imaging, but often involve expensive computation.
This paper proposes a new penalized weighted least aquares (PWLS) reconstruction approach that exploits regularization based on an efficient Union of Learned TRAnsforms (PWLS-ULTRA).
In the following, we briefly review recent methods for LDCT image reconstruction and summarize the contributions of this work.

%	\vspace{-0.1in}
\subsection{Background}     

Various methods have been proposed for image reconstruction in LDCT imaging.
When radiation dose is reduced, analytical filtered back-projection (FBP) image reconstruction methods (e.g., the Feldkamp-Davis-Kress or FDK method \cite{feldkamp:84:pcb}) typically provide unacceptable image quality. 
For example, streak artifacts increase severely as radiation dose is reduced \cite{imai:09:sco}. 
Model-based image reconstruction (MBIR) methods, aka statistical image reconstruction (SIR) methods, can provide high-quality reconstructions from low-dose scans \cite{fessler:00:sir, elbakri:02:sir}.
These methods iteratively find the image based on the system (physical) model, the measurement statistical model, and (assumed) prior information about the unknown object.
A typical MBIR method for CT uses a penalized weighted-least squares (PWLS) cost function with a statistically weighted quadratic data-fidelity term and a penalty term (regularizer) modeling prior knowledge of the underlying unknown object \cite{sauer:93:alu, thibault:06:arf, thibault:07:atd}.

Many current LDCT reconstruction methods use simple prior information. 
Adopting better image priors in MBIR could substantially improve image reconstruction quality for LDCT scans. 
The prior image constrained compressed sensing (PICCS) method was first proposed to enable accurate reconstruction of CT images from highly undersampled projection data sets \cite{chen:08:pic, ramirezgiraldo:11:npi, chen:12:tri}. 
Since a normal-dose CT image scanned previously may be available in some clinical applications, dose reduction using prior image constrained compressed sensing (DR-PICCS) was proposed to reduce image noise \cite{lauzier13:cos}.
Ma \mbox{\textit{et al.}} \cite{ma:11:ldc} proposed the previous normal-dose scan induced nonlocal means (ndiNLM) method to utilize the normal-dose image to enable low dose CT image reconstruction. 
The ndiNLM method expects that the normal-dose and the current low-dose scans are spatially aligned, and determines optimal local weights from the normal-dose image to improve the NLM weighted average \cite{ma:11:ldc, zhang:17:aon}. The PICCS and ndiNLM class of methods incorporate prior information from corresponding normal-dose CT images, assumed available. We propose a method that differs from these approaches in that it does not require prior normal-dose images of the same patient or object, and can rather learn general CT image features or filters from diverse image sets and datasets.

Extracting prior information from big datasets of CT images has great potential to enable MBIR methods to produce significantly improved reconstructions from LDCT measurements.  Images are often sparse in certain transform domains (such as wavelets, discrete cosine transform, and discrete gradient)  or dictionaries. The synthesis dictionary model approximates a signal by a linear combination of a few columns or atoms of a pre-specified dictionary \cite{bruckstein:09:fss}.
The choice of the synthesis dictionary is critical for the success of sparse representation modeling and other applications \cite{rubinstein:10:dfs}.
The data-driven adaptation of dictionaries, or dictionary learning \cite{olshausen:96:eos, engan:99:moo, aharon:06:ksa, yaghoobi:09:dlf, mairal:10:olf} yields dictionaries with better sparsifying capability for specific classes of signals than analytic dictionaries based on mathematical models.
Such learned dictionaries have been widely exploited in various applications in recent years, including super-resolution imaging, image or video denoising, classification, and medical image reconstruction \cite{elad:06:idv, mairal:07:srf, protter:09:isd, kong:12:adl, ravishankar:11:mir, chen:13:iat, lu:12:fvi}.
Some recent works also studied parametrized models such as adaptive tight frames \cite{zhou:13:atf}, multivariate Gaussian mixture distributions \cite{zhang:16:agm}, and shape dictionaries \cite{aghasi:13:ssr}.

Recently, Xu \textit{et al.} \cite{xu:12:ldx} applied dictionary learning to 2D LDCT image reconstruction by proposing a PWLS approach with an overcomplete synthesis dictionary-based regularizer. 
Their method uses either a global dictionary trained from 2D image patches extracted from a normal-dose FBP image, or an adaptive dictionary jointly estimated with the low-dose image. 
The trained global dictionary worked better than the adaptively estimated dictionary for highly limited (e.g., with very few views, or ultra-low dose) data. Several works proposed 3D CT reconstruction by learning either a 3D dictionary from 3D image patches, or learning three 2D dictionaries (dubbed 2.5D) from image patches extracted from slices along the x-y, y-z, and x-z directions, respectively \cite{liu:16:3fc, luo:16:25d}. 

Dictionary learning methods typically alternate between estimating the sparse coefficients of training signals or image patches (\emph{sparse coding step}) and updating the dictionary (\emph{dictionary update step}).
The sparse coding step in both synthesis dictionary learning \cite{elad:06:idv,aharon:06:ksa}  and analysis dictionary learning \cite{rubinstein:13:aka} is NP-Hard (Non-deterministic Polynomial-time hard) in general, and algorithms such as K-SVD \cite{elad:06:idv,aharon:06:ksa} involve relatively expensive computations for sparse coding. 
A recent generalized analysis dictionary learning approach called sparsifying transform learning \cite{ravishankar:13:lst,ravishankar:15:lst} more efficiently learns a transform model for signals. 
The transform model assumes that a signal $\x  \in \mathbb{R}^{n}$ is approximately sparsifiable using a transform $\omg  \in \mathbb{R}^{m\times n}$, i.e., $\omg \x = \z + \e$ where $\z \in  \mathbb{R}^{m}$ is sparse in some sense, and $\e \in  \mathbb{R}^{m}$ denotes the modeling error in the transform domain. 
Transform learning methods typically alternate between sparse approximation of training signals in the transform domain (\emph{sparse coding step}) and updating the transform operator (\emph{transform update step}). In contrast to dictionary learning methods, the sparse coding step in transform learning involves simple thresholding \cite{ravishankar:13:lst,ravishankar:15:lst}.
Transform learning methods have been recently demonstrated to work well in applications \cite{ravishankar:13:lds,wen:15:vdb,wen:17:fla,ravishankar:15:ebc}.
Pfister and Bresler \cite{pfister:14:mbi,pfister:14:trw,pfister:14:ast} showed the promise of PWLS reconstruction with adaptive square transform-based regularization, wherein they jointly estimated the square transform (ST) and the image. 
Pre-training a (global) transform from a large dataset would save computations during CT image reconstruction, and may also be well-suited for highly limited data (evidenced earlier for dictionary learning in \cite{xu:12:ldx}).

Wen \textit{et al. }recently extended the single ST learning method to learning a union of square transforms model, also referred to as an overcomplete transform with block cosparsity (OCTOBOS) \cite{wen:14:sos}.  
This transform learning approach jointly adapts a collection (or union) of $K$ square transforms and clusters the signals or image patches into $K$ groups. Each (learned) group of signals is well-matched to a corresponding transform in the collection. 
Such a learned union of transforms outperforms the ST model in applications such as image denoising \cite{wen:14:sos}. 

%\vspace{-0.1in}
\subsection{Contributions}     

Incorporating the efficient square transform (ST) model, we propose a new PWLS approach for LDCT reconstruction that exploits regularization based on a pre-learned square transform (PWLS-ST). We also extend this approach to a more general PWLS scheme involving a Union of Learned TRAnsforms (PWLS-ULTRA).
The transform models are pre-learned from numerous patches extracted from a dataset of CT images or volumes.  
We also incorporate patch-based weights in the proposed regularizer to help improve image resolution or noise uniformity.
We propose an efficient iterative algorithm for the PWLS costs that alternates between a \emph{sparse coding and clustering step} (which reduces to a \emph{sparse coding step} for PWLS-ST) that uses closed-form solutions, and an iterative \emph{image update step}.
There are several iterative algorithms that could be used for the image update step such as the preconditioned conjugate gradient (PCG) method \cite{fessler:99:cgp}, the separable quadratic surrogate method with ordered-subsets based acceleration (OS-SQS) \cite{erdogan:99:osa}, iterative coordinate descent (ICD) \cite{yu:11:fmb}, splitting-based algorithms \cite{ramani:12:asb}, and the optimal gradient method (OGM) \cite{kim:15:cos}. 
We chose the relaxed linearized augmented Lagrangian method with ordered-subsets (relaxed OS-LALM) \cite{nien:16:rla} for the image update step.

The proposed PWLS-ULTRA approach clusters the voxels into different groups. These groups often capture features such as bones, specific soft tissues, edges, etc.
Experiments with 2D and 3D axial CT scans of the extended cardiac-torso (XCAT) phantom and 3D helical chest and abdomen scans show that for both normal-dose and low-dose levels, the proposed methods significantly improve the quality of reconstructed images compared to conventional reconstruction methods such as filtered back-projection or PWLS reconstruction with a nonadaptive edge-preserving regularizer (PWLS-EP).
The union of learned transforms provides better image reconstruction quality than using a single learned square transform. The proposed PWLS-ULTRA achieves comparable or better image quality compared to learned overcomplete synthesis dictionaries, but importantly, is much faster (computationally more efficient).

We presented a brief study of PWLS-ST for low-dose fan-beam (2D) CT image reconstruction in \cite{zheng:16:ldc}. This paper investigates the more general PWLS-ULTRA framework, and presents experimental results illustrating the properties of the PWLS-ST and PWLS-ULTRA algorithms and  demonstrating their performance for low-dose fan-beam, cone-beam (3D) and helical (3D) CT.

%\vspace{-0.1in}
\subsection{Organization}      

Section~\ref{sec:formulation} describes the formulations for pre-learning a square transform or a union of transforms, and the formulations for PWLS reconstruction with regularization based on learned sparsifying transforms. 
Section~\ref{sec:algorithms} derives efficient optimization algorithms for the proposed problems.
Section~\ref{sec:results} presents experimental results illustrating properties of the proposed algorithms and demonstrating their promising performance for LDCT reconstruction compared to numerous recent methods.
Section~\ref{sec:conclusions} presents our conclusions and mentions areas of future work.

%\vspace{-0.1in}
\section{Problem Formulations for Transform Learning and Image Reconstruction}
\label{sec:formulation}

\subsection{PWLS-ST Formulation for LDCT Reconstruction}	

Given $N'$ vectorized image patches (2D or 3D) extracted from a dataset of CT images or volumes, we learn a square transform $\omg \in \mathbb{R}^{l \times l}$ by solving the following (training) optimization problem:
\begin{equation}\label{eq:P0}
	\min_{\omg,\Z}\| \omg\X-\Z \|_{F}^2 + \lambda Q(\omg)+ \sum_{i=1}^{N'} \eta^2\|\Z_{i}\|_0
	\tag{P0}
\end{equation}	
where $l$ is the number of pixels in each patch, $\lambda = \lambda_0 \|  \X\|_{F}^2 $ ($\lambda_0 >0$ is a constant) and $\eta > 0$ are scalar parameters, and $\{\Z_i\}_{i=1}^{N'}$ denote the sparse codes of the training signals (vectorized patches) $\{\X_i\}_{i=1}^{N'}$.  
Matrices $\X \in \mathbb{R}^{l \times N'} $ and $\Z \in \mathbb{R}^{l \times N'}$ have the training signals and sparse codes respectively, as their columns.
The $\ell_{0}$ ``norm'' counts the number of non-zeros in a vector.
The term $\| \omg\X-\Z \|_{F}^2$ is called the sparsification error and measures the deviation of the signals in the transform domain from their sparse approximations.
Regularizer  $Q(\omg) \triangleq \| \omg \|_{F}^2 - \log|\det \omg|$ prevents trivial solutions and controls the condition number of $\omg$ \cite{ravishankar:15:lst}.

After a transform $\omg $ is learned, we reconstruct an (vectorized) image or volume $\x \in \mathbb{R}^{N_p}$ from noisy sinogram data $\y \in \mathbb{R}^{N_d}$  by solving the following optimization problem \cite{zheng:16:ldc}: 
\begin{equation}\label{eq:P1}	
	\min_{\x \succeq \0}  \frac{1}{2}\|\y - \A \x\|^2_{\W}  + \beta \R(\x)   
	\tag{P1}
\end{equation}
where $\W = \diag \{w_i\} \in \mathbb{R}^{N_d \times N_d}$ is a diagonal weighting matrix with elements being the estimated inverse variance of $y_i$ \cite{thibault:06:arf}, $\A \in  \mathbb{R}^{  N_d  \times N_p}$ is the system matrix of a CT scan, the parameter $\beta>0$ controls the noise and resolution trade-off, and the regularizer $\R (\x)$ based on $\omg$ is defined as
\begin{equation}\label{eq:Rx_ST}
	\R(\x) \triangleq   \min_{\{\z_j\}}  \sum_{j=1}^{\tilde{N}} \tau_j  \bigg\{  \|\omg \P_j \x - \z_{j}\|^2_2 + \gamma^2\|\z_{j}\|_0 \bigg\}  
\end{equation}			
where $\tilde{N}$ is the number of image patches, the operator $\P_j \in \mathbb{R}^{l\times N_p}$ extracts the $j$th patch of $l$ voxels of $\x$ as $\P_j\x$, and vector $\z_{j} \in \mathbb{R}^{l}$ denotes the transform-sparse representation of $\P_j\x$. 
The regularizer includes a sparsification error term and a $\ell_{0}$ ``norm''-based sparsity penalty with weight $\gamma^{2}$ ($\gamma>0$).

We also include patch-based weights $\{\tau_j\}$ in \eqref{eq:Rx_ST} to encourage uniform spatial resolution or uniform noise in the reconstructed image \cite{chun:17:esv} as follows:
\begin{equation}\label{eq:tau_j}
	\tau_j  \triangleq  \|\P_j \kap\|_1 / l
\end{equation}			
with $\kap$ (of same size as $\x$) whose elements $\kappa_j$ are defined in terms of the entries of $\A$ (denoted $a_{ij}$) and $\W$ as $\kappa_j \triangleq \sqrt{{\sum_{i=1}^{N_d} a_{ij} w_i } / {\sum_{i=1}^{N_d}  a_{ij}}}$ \cite[eq(39)]{cho:15:rdf}.
While \eqref{eq:tau_j} uses the $\ell_{1}$ norm, corresponding to the mean value of $\P_j \kap$, to define $\tau_{j}$, we have observed that other alternative norms also work well in practice for LDCT reconstruction.

\begin{figure}[!t]
	\centering  	
	\includegraphics[width=0.5\textwidth]{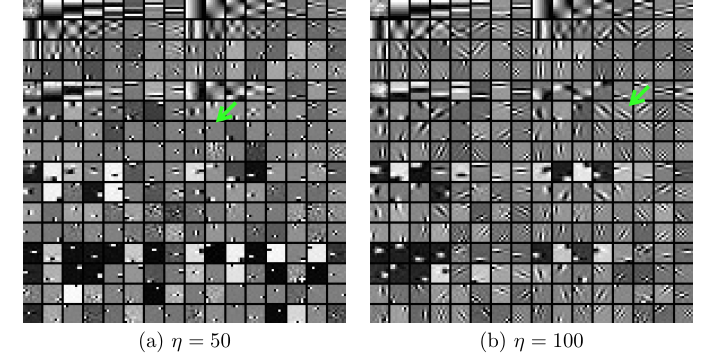}			
		\vspace{-0.25in}
	\caption{ Behavior of PWLS-ST: Pre-learned sparsifying transform $\omg$ with (a) $\eta =  50$ and (b) $\eta =  100$. The rows of the $512\times 512$ matrix $\omg$ are reshaped into $8\times 8 \times 8$ (3D) patches and the first $8\times 8$ slices of $256$ of these 3D patches are displayed for simplicity.} \label{fig:st_comp} 
	\vspace{-0.1in}
\end{figure}

Fig. \ref{fig:st_comp} shows example transforms (rows of $\omg$ are reshaped as $8 \times 8 \times 8$ patches and the first $8 \times 8$ slices of $256$ such 3D patches are shown) learned from $8 \times 8 \times 8$ patches of an XCAT phantom \cite{segars:08:rcs} volume. 
The transform learned with $\eta =100$ in \eqref{eq:P0} has more oriented features whereas the transform learned with $\eta=50$ shows more gradient (or finite-difference) type features (pointed by the green arrows).
This behavior suggests that a single ST may not be rich enough to capture the diverse features, edges, and other properties of CT volumes. 
Therefore, next we consider the extension of the ST approach to a rich union of learned transforms scheme.
%The next section extends the ST approach to a union of learned transforms scheme that outperforms the PWLS-ST approach.

%\vspace{-0.1in}
\subsection{Learning a Union of Sparsifying Transforms}

To learn a union of sparsifying transforms $\left \{ \omg_k \right \}_{k=1}^{K}$ from $N'$ (vectorized) patches, we solve
\begin{equation}\label{eq:P2}
	\begin{aligned}
		\min_{\{\omg_k,\Z_i,C_k\}}  &\sum_{k=1}^{K}    \sum_{i\in C_k}  \bigg\{  \| \omg_k\X_i-\Z _i\|_{2}^2 +  \eta^2\|\Z_{i}\|_0  \bigg\}  \\ &+ \sum_{k=1}^{K}   \lambda_k Q(\omg_k)  \quad \ \ \ \ \ \mathrm{s.t.} \ \ \{C_k\}\in \mathcal{G}.
	\end{aligned}
	\tag{P2}
\end{equation}		
This formulation groups the training signals $\{\X_i\}$ into $K$ classes according to the transform they best match, and $C_{k}$ denotes the set of indices of signals matched to the $k$th class. Set $\mathcal{G} $ denotes all possible partitionings of $\left \{ 1,2,..,N' \right \}$ into $K$ disjoint subsets.
We use $K$ regularizers $Q(\omg_k) \triangleq \| \omg_k \|_{F}^2 - \log|\det \omg_k|$, $1\leq k \leq K$, to control the properties of the transforms. 
We set these regularizer weights as $\lambda_k = \lambda_0 \|  \X_{C_k} \|_{F}^2 $ \cite{wen:14:sos}, where $\lambda_0 >0$ is a constant and $\X_{C_k}$ is a matrix whose columns are the training signals in the $k$th cluster. This choice of $\{\lambda_k\}$ together with $\eta = \eta_0  \|  \X \|_{F}$ for $\eta_0 >0$ allows the terms in \eqref{eq:P2} to scale appropriately with the data.
Problem \eqref{eq:P2} learns a collection of transforms and a clustering for the image patches, together with the patches' sparse coefficients $\{\Z_i\}$. The next section uses these transforms for image reconstruction.

\subsection{LDCT Reconstruction with ULTRA Regularization}

We propose a PWLS-ULTRA framework, where we solve \eqref{eq:P1} but with the regularizer $\R(\x)$ defined based on a union of sparsifying transforms as
\begin{equation}\label{eq:Rx_ULTRA}
	\begin{aligned}
		\R(\x) \triangleq  \min_{\{\z_j , C_k\} }   & \sum_{k=1}^{K}  \bigg\{   \sum_{j\in C_k}  \tau_j  \big\{ \|\omg_k \P_j \x - \z_{j}\|^2_2  +  \gamma^2\|\z_{j}\|_0  \big\}   \bigg\} \\  & \mathrm{s.t.}  \quad \{C_k\}\in \mathcal{G}.
	\end{aligned}
\end{equation}			
This regularizer measures the sparsification error of each patch using its best-matched transform. Using \eqref{eq:Rx_ULTRA}, \eqref{eq:P1} estimates the image $\x$, the sparse coefficients of image patches $\{\z_j\}$, and the cluster assignments $\{C_k\}$ from LDCT sinogram data $\y$.

%\vspace{-0.1in}
\section{Algorithms and Properties}
\label{sec:algorithms}
The square transform learning and the PWLS-ST formulations are special cases (corresponding to $K=1$) of the ULTRA-based formulations.
Therefore, this section describes algorithms for solving \eqref{eq:P1} with regularizer \eqref{eq:Rx_ULTRA} and \eqref{eq:P2}.

\vspace{-0.1in}
\subsection{Algorithm for Training a Union of Transforms} \label{subsec:learn_ultra}

We adopt an alternating minimization algorithm for \eqref{eq:P2} that alternates between a \textit{transform update step} (solving for $\{\omg_k\} $) and a \textit{sparse coding and clustering step} (solving for $\{\Z_i, C_k\}$). These steps are described next.

\subsubsection{ Transform Update Step}
With $\{\Z_i, C_k\}$ fixed, we solve the following optimization problem for $\{\omg_k\}$ \cite{wen:14:sos}:
\begin{equation}
	\min_{\{\omg_k\}}  \sum_{k=1}^{K}    \sum_{i\in C_k}   \| \omg_k\X_i-\Z _i\|_{2}^2 + \sum_{k=1}^{K}   \lambda_k Q(\omg_k).
\end{equation}		
Since the objective is in summation form, the above problem separates into $K$ independent single transform learning problems that we solve in parallel. The $k$th such optimization problem is as follows:
\begin{equation} \label{eq:single_ultra_learn}
	\min_{\omg_k}   \sum_{i\in C_k}  \| \omg_k\X_i-\Z _i\|_{2}^2  +  \lambda_k Q(\omg_k). 
\end{equation}		
We update the transform $\omg_k$ following prior work \cite{wen:14:sos,ravishankar:15:lst}.
Let $\QQ \SSS \RR^{T}$ denote the full singular value decomposition of $\LL^{-1} \X_{C_k} \Z_{C_k}^{T}$, with $\LL \LL^{T} \triangleq \X_{C_k} \X_{C_k}^{T} + \lambda_{k} \I$ (i.e., $\LL$ is a matrix square root). Then, the minimizer of \eqref{eq:single_ultra_learn} is
\begin{equation}\label{eq:omg}
	\hat{\omg}_k = 0.5 \RR\big({\SSS} +({\SSS}^2 +2\lambda_k\I)^{\frac{1}{2}}\big) \QQ^{T} \LL^{-1}.
\end{equation} 

\subsubsection{Sparse Coding and Clustering Step}

With $\{\omg_k\}$ fixed, we solve the following sub-problem for $\{\Z_i, C_k\}$:
\begin{equation}\label{eq:learn_spar}
	\min_{\{\Z_i,C_k\}}  \sum_{k=1}^{K}    \sum_{i\in C_k}  \bigg\{  \| \omg_k\X_i-\Z _i\|_{2}^2 +  \eta^2\|\Z_{i}\|_0  + \lambda_0 \|  \X_{i} \|_{2}^2  Q(\omg_k)  \bigg\} 
\end{equation}		
For given cluster memberships, the optimal sparse codes are ${\Z}_{i} =  H_{\eta}(\omg_k \X_i) , \forall i \in C_k$, where the \textit{hard-thresholding} operator $H_{\eta}(\cdot)$ zeros out vector entries with magnitude less than $\eta$.
Using this result, it follows that the optimal cluster membership for each $\X_i$ in \eqref{eq:learn_spar} is $\hat{k}_i = \underset{1\leq k\leq K}{\operatorname{arg\,min}} \big\{  \|\omg_k \X_i - H_{\eta}(\omg_k \X_i)\|^2_2 + \eta^2\|H_{\eta}(\omg_k \X_i)\|_0 + \lambda_0 \|  \X_{i} \|_{2}^2  Q(\omg_k)  \big\}$, and the corresponding optimal sparse code is $\hat{\Z}_{i} =  H_{\eta}(\omg_{\hat{k}_i} \X_i)$.

%\vspace{-0.1in}
\subsection{PWLS-ULTRA Image Reconstruction Algorithm}

We propose an alternating algorithm for the PWLS-ULTRA formulation (i.e., \eqref{eq:P1} with regularizer \eqref{eq:Rx_ULTRA}) that alternates between updating $\x$ (\textit{image update step}),  and $\{\z_j, C_k\}$ (\textit{sparse coding and clustering step}).  	

\subsubsection{Image Update Step} \label{subsub:image_update}

With $\{\z_j, C_k\}$ fixed, \eqref{eq:P1} for PWLS-ULTRA reduces to the following weighted least squares problem:
\begin{equation}
	\label{eq:image}
	\min_{\x \succeq \0} \frac{1}{2} \|\y - \A\x \|^2_{\W} + \R_2(\x)  
\end{equation}
where $\R_2(\x) \triangleq \beta \sum_{k=1}^{K} \sum_{j\in C_k} \tau_j  \|\omg_k \P_j \x - \z_{j}\|^2_2  $.

We solve \eqref{eq:image} using the recent relaxed OS-LALM \cite{nien:16:rla}, whose iterations are shown in Algorithm \ref{alg: ultra}.  Here, for each iteration $n$,  we further iterate over $1\leq m \leq M$ corresponding to $M$ ordered subsets.
The matrices $\A_m$, $\W_m$, and the vector $\y_m$ in Algorithm \ref{alg: ultra} are sub-matrices  of $\A$, $\W$, and sub-vector of $\y$, respectively, for the $m$th subset.
Matrix $\D_{\A} \succeq \A^{T}\W\A$ is a diagonal majorizing matrix of $\A^T\W\A$; specifically we use \cite{erdogan:99:osa} 
\begin{equation}\label{eq:DA}		
	\D_{\A} \triangleq \diag \{\A^{T}\W\A\mathbf{1}\}\succeq \A^{T}\W\A.
\end{equation}		
The gradient $\nabla \R_2(\x)= 2 \beta \sum_{k=1}^{K} \sum_{j\in C_k} \tau_j  \P_j^T \omg_k^T (\omg_k \P_j \x - \z_{j})$, the (over-)relaxation parameter $\alpha \in [1, 2)$, and the parameter $\rho >0 $ decreases gradually with iteration \cite{nien:16:rla}, 
\begin{equation}\label{eq:rho}		
	\rho_r(\alpha)= \begin{cases}
		1,      &  r=0\\ 
		\frac{\pi}{\alpha(r+1)}\sqrt{1-\big(\frac{\pi}{2\alpha(r+1)}\big)^2},     &  \text{otherwise,}
	\end{cases}
\end{equation}		
where $r$ indexes the total number of $n$ and $m$ iterations.
Lastly, $\D_\R$ in Algorithm \ref{alg: ultra}  is a diagonal majorizing matrix of the Hessian of the regularizer $\R_2(\x)$, specifically:
\begin{equation}\label{eq:DR}		
	\vspace{-0.02in}
	\begin{aligned}
		\D_\R  &\triangleq 2 \beta \bigg\{\max_k \lambda_{\max}(\omg_k^{T} \omg_k) \bigg\} \sum_{k=1}^{K}  \sum_{j\in C_k}\tau_j \P_j^{T}\P_j \\
		%		& \succeq   2 \beta  \sum_{k=1}^{K}   \bigg\{ \lambda_{\max}(\omg_k^{T} \omg_k)   \sum_{j\in C_k} \tau_j \P_j^{T}\P_j \bigg\}  \\
		& \succeq  2\beta \sum_{k=1}^{K} \sum_{j\in C_k} \tau_j \P_j^{T} \omg_k^{T} \omg_k \P_j = \nabla^2 \R_2(\x).
	\end{aligned}
	\vspace{-0.02in}
\end{equation}
Since this $\D_\R$ is independent of $\x$,  $\{\z_j\}$, and $\{C_k\}$,  we precompute it using patch-based operations \cite{ravishankar:11:mir} (cf. the supplement\footnote{Supplementary material is available in the supplementary files/multimedia tab.} for details) prior to iterating.

\subsubsection{Sparse Coding and Clustering Step}

With $\x$ fixed,  we solve the following sub-problem to determine the optimal sparse codes and cluster assignments for each patch:
\begin{equation}
	\min_{\{\z_{j} \}, \{C_k\}\in \mathcal{G}  }  \sum_{k=1}^{K} \bigg\{   \sum_{j\in C_k} \tau_j  \big\{   \|\omg_k \P_j \x - \z_{j}\|^2_2 + \gamma^2\|\z_{j}\|_0 \big\} \bigg\}.    
\end{equation} 

%\vspace{-0.1in}
For each patch $\P_j \x$, with (optimized) ${\z}_{j} =  H_{\gamma}(\omg_k \P_j \x)$, the optimal cluster assignment is computed as follows:
\begin{equation}\label{eq:k_j}
	\hat{k}_j = \underset{1\leq k\leq K}{\operatorname{arg\,min}}  \|\omg_k \P_j \x - H_{\gamma}(\omg_k \P_j \x)\|^2_2 + \gamma^2  \| H_{\gamma}(\omg_k \P_j \x) \|_0
\end{equation}	
Minimizing over $k$ above finds the best-matched transform. Then, the optimal sparse codes are $\hat{\z}_{j} =  H_{\gamma} (\omg_{\hat{k}_j} \P_j \x)$.

\subsubsection{Overall Algorithm}

The proposed method for the PWLS-ULTRA problem is shown in Algorithm \ref{alg: ultra}. The algorithm for the PWLS-ST formulation is obtained by setting $K=1$ and skipping the clustering procedure in the sparse coding and clustering step.
Algorithm \ref{alg: ultra} uses an initial image estimate and the union of pre-learned transforms $\{\omg_k\}$. It then alternates between the image update, and sparse coding and clustering steps until a convergence criterion (such as  $ \| \tilde{\x}^{(t+1)} -  \tilde{\x}^{(t)}\|_{2}<\epsilon$ for some small $\epsilon>0$) is satisfied, or alternatively until some maximum inner/outer iteration counts are reached.

\begin{algorithm}[!t]  
	\caption{PWLS-ULTRA Algorithm}\label{alg: ultra}
	\begin{algorithmic}[0]
		\State \textbf{Input:}
		initial image $\tilde{\x}^{(0)}$, pre-learned $\{\omg_k\}$, threshold $\gamma$, $\alpha = 1.999$,
		$\D_{\A} $ in \eqref{eq:DA}, $\D_\R $ in \eqref{eq:DR}, number of outer iterations $T$, number of inner iterations $N$, and number of subsets $M$.											
		\State \textbf{Output:}  reconstructed image $\tilde{\x}^{(T)}$, cluster indices $\{\tilde{C}_k^{(T)}\}$.
		\For {$t =0,1,2,\cdots,{T-1}$}		
		
		\State \textbf{1) Image Update}: $\{ \tilde{\z}_j^{(t)}\}$ and  $\{\tilde{C}_k^{(t)}\}$ fixed,	
		
		\textbf{Initialization:} $\rho=1$, $\x^{(0)} = \tilde{\x}^{(t)}$, $\g^{(0)} = \ze^{(0)}  = M\A_M^{T}\W_M(\A_M\x^{(0)}-\y_M) $ and $\h^{(0)} = \D_\A \x^{(0)} - \ze^{(0)}$.

		\For {$n =0,1,2,\cdots,N-1$}	
		\For {$m =0,1,2,\cdots,M-1$}  $r = nM +m$			
		\begin{equation*}
			\left\{			
			\begin{aligned}
				\s^{(r+1)} &= \rho(\D_\A \x^{(r)} -\h^{(r)}) + (1-\rho)\g^{(r)} \\
				\x^{(r+1)} &= [\x^{(r)} - (\rho\D_\A+\D_\R)^{-1}(\s^{(r+1)} +\nabla \R_2(\x^{(r)}))]_+ \\
				\ze^{(r+1)} & \triangleq  M \A^T_m\W_m(\A_m\x^{(r+1)}-\y_m)   \\
				\g^{(r+1)} &= \frac{\rho}{\rho+1}(\alpha \ze^{(r+1)} + (1-\alpha)\g^{(r)}) +  \frac{1}{\rho+1}\g^{(r)}\\
				\h^{(r+1)}  &= \alpha(\D_{\A} \x^{(r+1)} -\ze^{(r+1)}) + (1-\alpha)\h^{(r)} 
			\end{aligned}
			\right.
		\end{equation*}  
		\State decreasing $\rho$ using \eqref{eq:rho}. % where $r = nM +m$.		 
		\EndFor			
		\EndFor	
		\State   $\tilde{\x}^{(t+1)} = \x^{(NM)}$. 
		\State \textbf{2) Sparse Coding and Clustering}: with $\tilde{\x}^{(t+1)}$ fixed, for each $1\leq j \leq N$, obtain $\hat{k}_j$ using \eqref{eq:k_j}. Then $\tilde{\z}_{j}^{(t+1)} = H_{\gamma}(\omg_{\hat{k}_j} \P_j \tilde{\x}^{(t+1)}) $, and update $\tilde{C}_{\hat{k}_j}^{(t+1)}$.
		\EndFor		
	\end{algorithmic}
\end{algorithm}

%\vspace{-0.1in}
\subsubsection{Computational Cost}

Each outer iteration of the proposed Algorithm \ref{alg: ultra} involves the image update and the sparse coding and clustering steps. The cost of the sparse coding and clustering step scales as $O(l^2N)$ and is dominated by matrix-vector products.
Importantly, unlike prior dictionary learning-based works \cite{xu:12:ldx}, where the computations for the sparse coding step (involving orthogonal matching pursuit (OMP) \cite{pati:93:omp}) can scale worse as $O(l^{3}N)$ (assuming synthesis sparsity levels of patches $\propto l$), the exact sparse coding and clustering in PWLS-ULTRA is cheaper, especially for large patch sizes.
Similar to prior works \cite{xu:12:ldx}, the computations in the image update step are dominated by the forward and back projection operations.  
Section \ref{sec:results} compares the proposed method to synthesis dictionary learning-based approaches, and shows that our transform approach runs much faster.

\section{Experimental Results}
\label{sec:results}
This section presents experimental results illustrating properties of the proposed algorithms and demonstrating their promising performance for LDCT reconstruction compared to numerous recent methods. We include additional experimental results in the supplement. A link to software to reproduce our results is provided at \mbox{\url{http://web.eecs.umich.edu/~fessler/irt/reproduce/}}.

\vspace{-0.1in}
\subsection{Framework and Data} \label{sec:resultsa}

We evaluate the proposed PWLS-ULTRA and PWLS-ST (i.e., with $K=1$) methods for 2D fan-beam and 3D axial cone-beam CT reconstruction of the XCAT phantom \cite{segars:08:rcs}. We also apply the proposed methods to helical CT clinical data of the chest and abdomen.

Section~\ref{sec:para_selection} discusses the role and intuition of each parameter in the proposed methods. Section~\ref{sec:resultsb} illustrates the properties of the transform learning and image reconstruction methods.
Sections~\ref{sec:resultsc} and \ref{sec:resultsd} show results for 2D fan-beam and 3D axial cone-beam CT, respectively, for the XCAT phantom data.
We used the ``Poisson + Gaussian'' model, i.e., $\tilde{k} \, \text{Poisson}\{I_0 \exp(-[\A\x]_i)\} + \text{Normal}\{0, \sigma^2\}$ to simulate CT measurements of the XCAT phantom, where $I_0$ is the incident X-ray intensity incorporating X-ray source illumination and the detector gain, the parameter $\tilde{k}=1$ models the conversion gain from X-ray photons to electrons, and $\sigma^2 = 5^2$ is the variance of electronic noise \cite{ding:16:mmp}. 
We compare the image reconstruction quality obtained with PWLS-ST and PWLS-ULTRA with those of:
\begin{itemize}
	\item{\textbf{FBP}: conventional FBP method with a Hanning window. }
	\item{\textbf{PWLS-EP}: PWLS reconstruction with the edge-preserving regularizer $\R(\x) = \sum_{j  =1}^{N_p} \sum_{k\in N_{j}}\kappa_{j} \kappa_{k} \varphi(x_j - x_k)$, where $N_j$ is the size of the neighborhood, $\kappa_j$ and $\kappa_k$ are the parameters encouraging uniform noise \cite{cho:15:rdf}, and 
		${\varphi}{(t)}\triangleq\delta^2 (  | t/\delta | - \log(1+| t/\delta |) )$. We optimized this PWLS cost function using the relaxed OS-LALM \cite{nien:16:rla}. }
	\item{\textbf{PWLS-DL}: PWLS reconstruction with a learned overcomplete synthesis dictionary based regularization, whose image update step is optimized by relaxed OS-LALM instead of the SQS-OS used in \cite{xu:12:ldx}. }
\end{itemize}
Section~\ref{sec:resultse} reports the reconstructions from helical CT clinical data of the chest and abdomen (low-dose). Finally, Section \ref{sec:resultsf} compares the performance of PWLS-ULTRA to an oracle scheme that uses cluster memberships estimated directly from the reference or ground truth images.

To compare various methods quantitatively for the case of the XCAT phantom, we calculated the Root Mean Square Error (RMSE) and Structural Similarity Index Measurement (SSIM) \cite{wang:04:iqa} of the reconstructions in a region of interest (ROI). 
RMSE in Hounsfield units \footnote{Modified Hounsfield units, where air is $0$ HU and water is $1000$ HU.} (HU) is defined as \mbox{RMSE $= \sqrt{\sum_{i=1}^{N_{p, \text{ROI}}}(\hat{x}_i-x^*_i)^2/{N_{p, \text{ROI}}}}$}, where $\x^*$ is the ground truth image and $N_{p, \text{ROI}}$ is the number of pixels (voxels) in the ROI. Unless otherwise noted, we tuned the parameters of various methods for each experiment to achieve good RMSE and SSIM.
For the clinical chest and low-dose abdomen data, the reconstructions were evaluated visually using voxel profiles. We display all reconstructions in this section using a display window $[800, 1200]$ HU, unless otherwise noted.

In the 2D fan-beam CT experiments, we pre-learned square transforms and union of square transforms from $8 \times 8$ overlapping image patches extracted from five $512 \times 512$ XCAT phantom slices, with a patch stride $1\times1$. We ran $1000$ iterations of the alternating minimization transform learning algorithm in Section \ref{subsec:learn_ultra} (or in \cite{ravishankar:15:lst} when $K=1$) to ensure convergence, and used $\lambda_0= 31$. The transforms were initialized with the 2D DCT, and k-means clustering (of patches) was used to initialize the clusters for learning a union of transforms.
We simulated a 2D fan-beam CT scan using an $840 \times 840$ XCAT phantom slice (air cropped) that differs from the training slices, and $\Delta_x=\Delta_y=0.4883$ mm. Noisy sinograms of size $888 \times 984 $ were numerically simulated with GE LightSpeed fan-beam geometry corresponding to a monoenergetic source with $1\times10^4$ and $5\times10^3$ incident photons per ray and no scatter, respectively. We reconstructed a $420 \times 420$ image with a coarser grid, where $\Delta_x=\Delta_y=0.9766$ mm. 
The ROI here was a circular (around center) region containing all the phantom tissues.

In the 3D cone-beam CT reconstruction experiments, we pre-learned STs and union of square transforms from $8 \times 8 \times 8$ patches ($N' \approx 1 \times 10^6$) extracted from a $420 \times 420 \times 54$ XCAT phantom (air cropped) with a patch stride $2 \times 2 \times 2$. We set $\lambda_0$ large enough, e.g., $\lambda_0= 31$, to ensure well-conditioned learned transforms. We ran the alternating minimization transform learning algorithms for $1000$ iterations. The transforms were initialized with the 3D DCT, and a random initialization was used for the clusters (because k-means produced some empty clusters for large $K$) for learning a union of square transforms.
We simulated an axial cone-beam CT scan using an $840 \times 840 \times 96$ XCAT phantom with $\Delta_x=\Delta_y=0.4883$ mm and $\Delta_z=0.625$ mm. 
We generated sinograms of size $888 \times 64 \times 984 $ using GE LightSpeed cone-beam geometry corresponding to a monoenergetic source with $1 \times 10^4$ and $5\times10^3$ incident photons per ray and no scatter, respectively. 
We reconstructed a $420 \times 420 \times 96$ volume with a coarser grid, where $\Delta_x=\Delta_y=0.9766$ mm and $\Delta_z=0.625$ mm.
For PWLS-ST and PWLS-ULTRA reconstructions, the patch size was $8 \times 8 \times 8$ with a patch stride $2 \times 2 \times 2$ ($\tilde{N} \approx  2 \times 10^6$ patches). 
The ROI for the 3D case consisted of the central $64$ of $96$ axial slices and a circular (around center) region in each slice (cylinder in 3D). The diameter of the circle was $420$ pixels, which is the width of each slice.

For the clinical chest data, we reconstructed a $420\times 420\times 222$ image volume (air cropped) with patch size $8 \times 8 \times 8$ and patch stride $3 \times 3 \times 3$ ($\tilde{N} \approx  1.5 \times 10^6$ patches), where $\Delta_x=\Delta_y=1.1667$ mm and $\Delta_z=0.625$ mm, from a helical CT scan. The size of the sinogram was $888\times 64\times 3611$ and pitch was $1.0$ (about $3.7$ rotations with rotation time $0.4$ seconds). The tube current and tube voltage of the X-ray source were $750$ mA and $120$ kVp, respectively.  To further evaluate the proposed method, we reconstructed $512\times 512\times 200$ abdomen region volumes with patch size $8 \times 8 \times 8$, patch stride $3 \times 3 \times 3$, $\Delta_x=\Delta_y=1$ mm and $\Delta_z=0.625$ mm, from low-dose helical CT patient scans. The size of the sinogram was $888\times 64\times 2952$ and pitch was $1.375$ ($3$ rotations with rotation time $0.8$ seconds). The tube voltage was $120$ kVp, and the tube currents were $150$ mA and $35$ mA (scanned twice for the same patient).

%\vspace{-0.1in}

\subsection{Parameter Selection}  \label{sec:para_selection}
The $\{\tau_j\}$ parameters are designed using the $\kap$ information as per \eqref{eq:tau_j}, so no additional tuning is needed.
Since the transforms are pre-learned once from a given dataset and used to reconstruct new data, the parameters $\lambda$ and $\eta$ are tuned during training. 
As mentioned in prior work \cite{ravishankar:15:lst}, the parameter $\lambda$ controls the condition number and larger values of $\lambda$ encourage well-conditioned transforms that work well for image reconstruction. 
The $\eta$ parameter can be set to achieve low sparsity (e.g., $5-10\%$) and a good trade-off with sparsification error (the transform-domain residual in the training objective) for training data. 
In our experiments, we learned transforms for a couple different $\eta$ values (training sparsities) and compared their effectiveness in some test reconstructions before picking the best learned model.

During reconstruction, mainly the parameters $\beta$ and  $\gamma$ (Section \ref{sec:resultsb} discusses about $K$) need to be tuned. These parameters are tuned to achieve a good trade-off between image resolution and noise. 
For example, large values of $\gamma$ would achieve very low sparsities and reduce the noise but potentially oversmooth the image. 
For a given learned transform, we tuned $\beta$ and $\gamma$ together to achieve good RMSE and SSIM of the reconstruction. 
Since the PWLS-ST and PWLS-ULTRA formulations are quite similar, except for the richer model and implicit clustering in the latter case, one could tune $\beta$ and $\gamma$ for ST first, and use these optimized values for ULTRA. 
In our experiments, we tuned parameters separately for ST and ULTRA, and found the tuned values to be typically similar.

Likewise, standard methods like the PWLS-EP method have an overall regularization parameter $\beta$ and an edge-preserving parameter $\delta$, so the number of parameters that one must tune during reconstruction (after training is done) is similar for EP and ULTRA. 
Similarly as for PWLS-ULTRA, the parameters (maximum patch-wise sparsity level and error threshold for sparse coding) for the prior PWLS-DL were selected carefully (by sweeping over values in a grid) to achieve good RMSE and SSIM in each case, for fair comparison.

%\vspace{-0.1in}
\subsection{Behavior of the Learning and PWLS-ULTRA Algorithms} \label{sec:resultsb}

\begin{figure}[!t]
	\centering  	
	\includegraphics[width=0.49\textwidth]{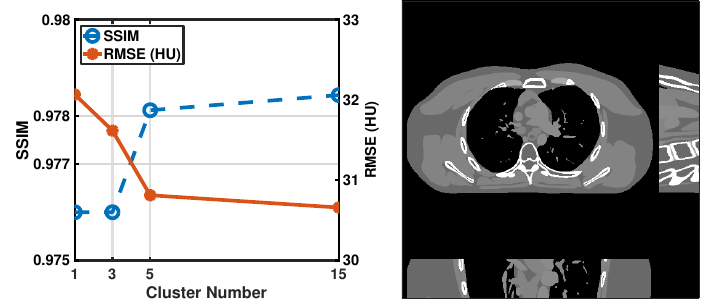}			
	\vspace{-0.25in}
	\caption{RMSE and SSIM for PWLS-ULTRA for various choices of number of clusters $K$ (left), and the central slices along three directions for the underlying volume in the cone-beam CT reconstruction experiments (right).}
	\label{fig:comp_conv}
	\vspace{-0.15in}
\end{figure}

We evaluate the behavior of the PWLS-ULTRA method (with $\tau_j =1 \, \forall j$) for 3D cone-beam CT data with $I_0 = 1\times10^4$. 
Fig.~\ref{fig:comp_conv} shows the central slices along three directions for the underlying (true) XCAT phantom volume. We reconstruct the volume from low-dose CT measurements.
Fig.~\ref{fig:comp_conv} shows the RMSE and SSIM of PWLS-ULTRA for various choices of $K$, the number of clusters (patch size $8 \times 8 \times 8$ and patch stride $2 \times 2 \times 2$). 
Rich models (large $K$) produce better reconstructions compared to using a single ST ($K=1$).
For the piece-wise constant phantom, $K=5$ clusters works well enough, with only a small additional RMSE or SSIM improvement observed for larger $K$. Larger values of $K$ led to sharper image edges.

\begin{figure*}[!t]
	\centering
	\includegraphics[width=1\textwidth]{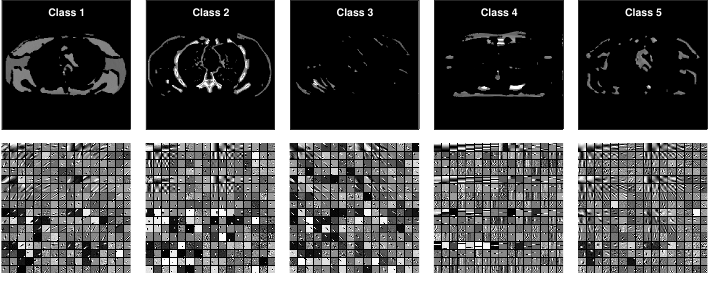}			
	\vspace{-0.35in}
	\caption{Pixel-level clustering results (top row) for the central axial slice of the PWLS-ULTRA  ($K = 5$) reconstruction at $I_0 = 1\times 10^4$. The pixels in each class are displayed using the intensities in the reconstruction.  The corresponding transforms (the first $8 \times 8$ slice of $8 \times 8 \times 8$ atoms) are in the bottom row. 
	}
	\label{fig:clusters}
	\vspace{-0.2in}
\end{figure*}

Fig.~\ref{fig:clusters} presents an example of the pixel-level clustering in the central axial slice achieved with the PWLS-ULTRA method for $K = 5$.
Since PWLS-ULTRA clusters patches, we cluster individual pixels using a majority vote among the 3D patches that overlap the pixel.
Class $1$ contains most of the soft tissues; class $2$ comprises most of the bones and blood vessels; classes $3$ and $4$ have some high-contrast edges oriented along specific directions; and class $5$ mainly includes low-contrast edges.
Since the clustering step (during both training and reconstruction) is unsupervised, i.e., different anatomical structures were not labeled manually, there are also a few edges with high pixel intensities included in class $2$.
The trained (3D) transforms (with $\eta = 50$) for each cluster are also displayed in a similar manner as in Fig.~\ref{fig:st_comp}. The transforms show features (e.g., with specific orientations) that clearly reflect the properties of the patches/tissues in each class.

%\vspace{-0.1in}
\subsection{2D LDCT Reconstruction Results and Comparisons} \label{sec:resultsc}

\subsubsection{Reconstruction Quality} \label{subsec:2d_comp}

We evaluate the performance of various algorithms for image reconstruction from low-dose fan-beam CT data.
Initialized with FBP reconstructions, we ran the PWLS-EP algorithm for $50$ iterations using relaxed OS-LALM with $24$ subsets, and set $\delta=10$ (HU) and the regularization parameter $\beta = 2^{16.0}$ and $\beta = 2^{16.5}$ for $I_0 = 1 \times 10^4$ and $I_0 = 5 \times 10^3$, respectively. 
For PWLS-DL, PWLS-ST, and PWLS-ULTRA, we initialized with the PWLS-EP reconstruction, and ran $200$ outer iterations with $2$ iterations of the image update step with $4$ ordered subsets, i.e., $N = 2$, $M = 4$.
For PWLS-DL, we pre-learned a $64 \times 256$ overcomplete dictionary from $8 \times 8$ patches extracted from five XCAT phantom slices (same slices as used for transform learning) with a patch stride $1\times 1$, using a maximum patch-wise sparsity level of $20$ and an error threshold or tolerance for sparse coding of $10^{-1}$.
During reconstruction with PWLS-DL, we used a maximum sparsity level of $25$, an error tolerance of $55$, and a regularization parameter of $7.0\times 10^4$ and $6.0\times 10^4$ for $I_0 = 1 \times 10^4$ and $I_0 = 5 \times 10^3$, respectively.
For PWLS-ST and PWLS-ULTRA ($K=15$), we chose  $\left ( \beta, \gamma, \eta \right )$ for the two incident photon intensities as follows:  $\left ( 2.0 \times 10^5, 20, 75 \right )$ and $\left ( 1.3 \times 10^5, 20, 75 \right )$ for PWLS-ST ($\tau_j = 1$); $\left ( 2.0 \times 10^5, 20, 125  \right )$ and $\left (   1.0 \times 10^5, 25, 125 \right )$ for PWLS-ULTRA ($\tau_j = 1$), and $\left ( 1.3 \times 10^4, 22, 125  \right )$ and $\left (   1.0 \times 10^4, 25, 125 \right )$ for PWLS-ULTRA with the weights $\tau_j$.

\begin{table}[!t]	
	\centering
	\caption{RMSE (HU) and SSIM of 2D (fan-beam) image reconstructions with FBP, PWLS-EP, PWLS-DL, PWLS-ST,  PWLS-ULTRA ($K=15$), and PWLS-ULTRA ($K=15$) with patch-based weights ($\tau_{j}$), for two incident photon intensities}
	\label{tab:2d}	 	
	\vspace{-0.1in}
	\begin{tabular}{ccccccc}			
		\toprule
		Intensity                                         &    FBP         &  EP           &  DL         & ST        & ULTRA   & ULTRA-$\{\tau_j\}$   \\
		\midrule
		\multirow{2}{*}  {$1\times10^4$}    &73.7          & 39.4       &  33.6       & 36.5      & 34.4     & \bf{33.1}    \\
		\cmidrule{2-7}                                &0.547        & 0.892       &    0.966    & 0.966    & 0.967   & \bf{0.969}    \\
		\midrule
		\multirow{2}{*}  {$5\times10^3$}    &89.0          & 49.7      &  39.1              & 43.9       & 39.8  & \bf{38.9}    \\
		\cmidrule{2-7}                                &0.472         & 0.884        &  \bf{0.958}  &  0.955    & 0.953 & 0.956    \\
		\bottomrule
	\end{tabular}
	\vspace{-0.15in}
\end{table}

\begin{table}[!t]	
	\centering
	\caption{RMSE (HU) in three ROIs of 2D (fan-beam) image reconstructions with FBP, PWLS-EP, PWLS-DL, PWLS-ST,  PWLS-ULTRA ($K=15$), and PWLS-ULTRA ($K=15$) with patch-based weights ($\tau_{j}$), for two incident photon intensities}
	\label{tab:2d_roi}	 	
	\vspace{-0.1in}
	
	\begin{tabular}{ccccc}			
		\toprule
		Intensity                                         & Methods                                   &  ROI-$1$           &  ROI-$2$          & ROI-$3$         \\
		\midrule
		\multirow{8}{*}  {$1\times10^4$}    &FBP                           &  21.8           &    15.6           &   39.6    \\
		\cmidrule{2-5}                                  &EP                             &  6.6            &    10.9           &   14.7    \\
		\cmidrule{2-5}                                 &DL                             &  \bf{3.7}            &    9.9      &   16.6    \\
		\cmidrule{2-5}                                 &ST                             &  3.9            &    10.8           &   14.1    \\
		\cmidrule{2-5}                               &ULTRA                          & 4.2             &  9.6              &     13.8    \\
		\cmidrule{2-5}                                &ULTRA-$\{\tau_j\}$      & 4.2           & \bf{9.3}          &     \bf{12.6}    \\
		
		\midrule
		\midrule
		\multirow{9}{*}  {$5\times10^3$}    &FBP                             &  51.7            &    36.4           &   39.0    \\
		\cmidrule{2-5}                                &EP                                & 7.1              & 14.9                &  28.5  \\
		\cmidrule{2-5}                                &DL                                & 7.0              & 14.5                &  20.7  \\
		\cmidrule{2-5}                               &ST                               & 6.3                & 14.3                &   21.4    \\
		\cmidrule{2-5}                               &ULTRA                          & \bf{ 5.8 }     &  \bf{ 13.7 }         &     \bf{17.5}    \\
		\cmidrule{2-5}                              &ULTRA-$\{\tau_j\}$      &  5.9            & \bf{ 13.7 }           &   18.1  \\
		\bottomrule
	\end{tabular}
	\vspace{-0.2in}
\end{table}

\begin{figure}[!h]
	%	\vspace{-0.05in}
	\centering  	
	\begin{tikzpicture}
	[spy using outlines={rectangle,green,magnification=2,size=10mm, connect spies}]
	\node {\includegraphics[width=0.225\textwidth]{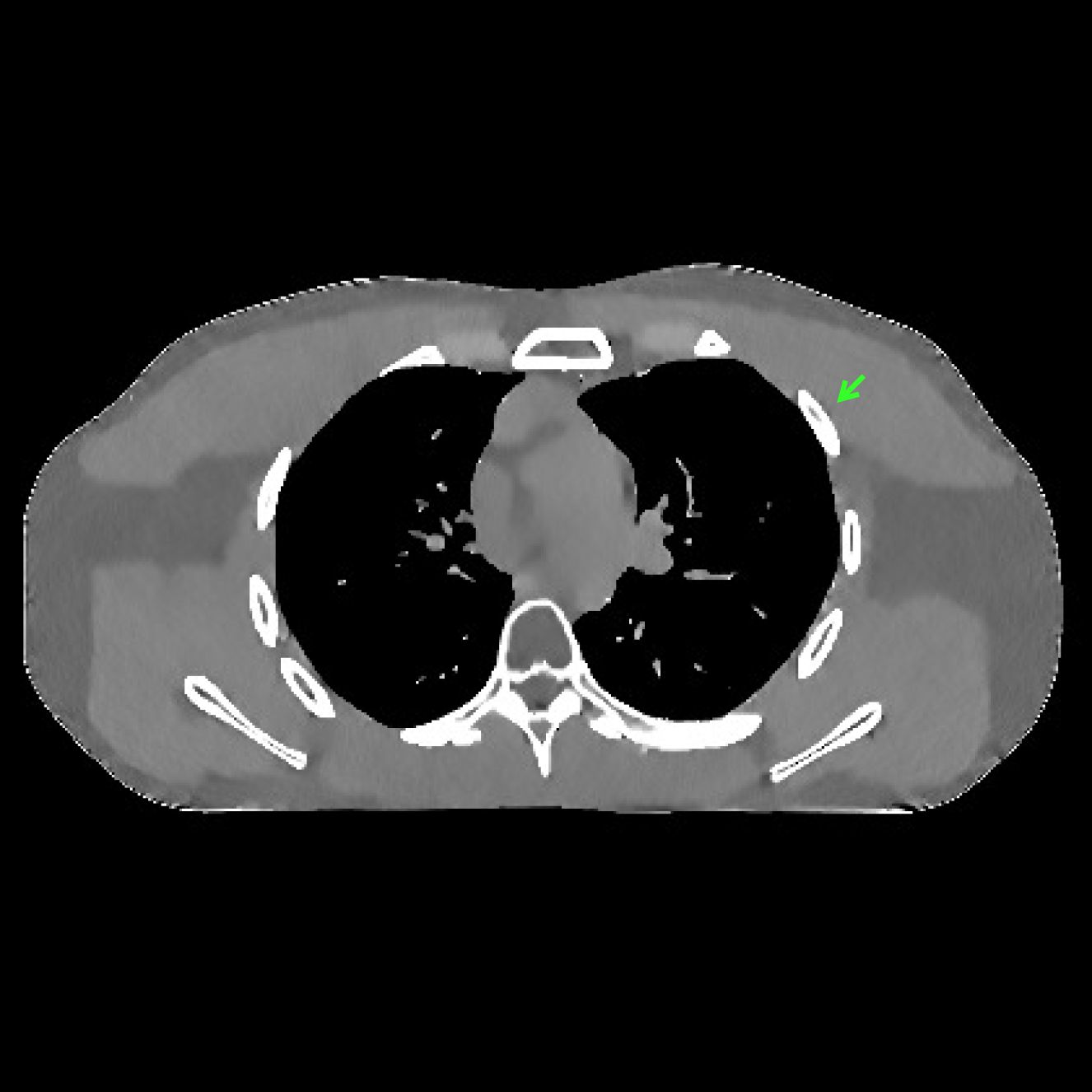}	};
	\spy on (-1.6,-0.2) in node [left] at (-0.9,1.5);		
	\spy on (0,0.25) in node [left] at (1,-1.45);	
	\spy on (1.1,0.5) in node [left] at (1.9,1.5);	
	\end{tikzpicture}
	\begin{tikzpicture}
	[spy using outlines={rectangle,green,magnification=2,size=10mm, connect spies}]
	\node {\includegraphics[width=0.225\textwidth]{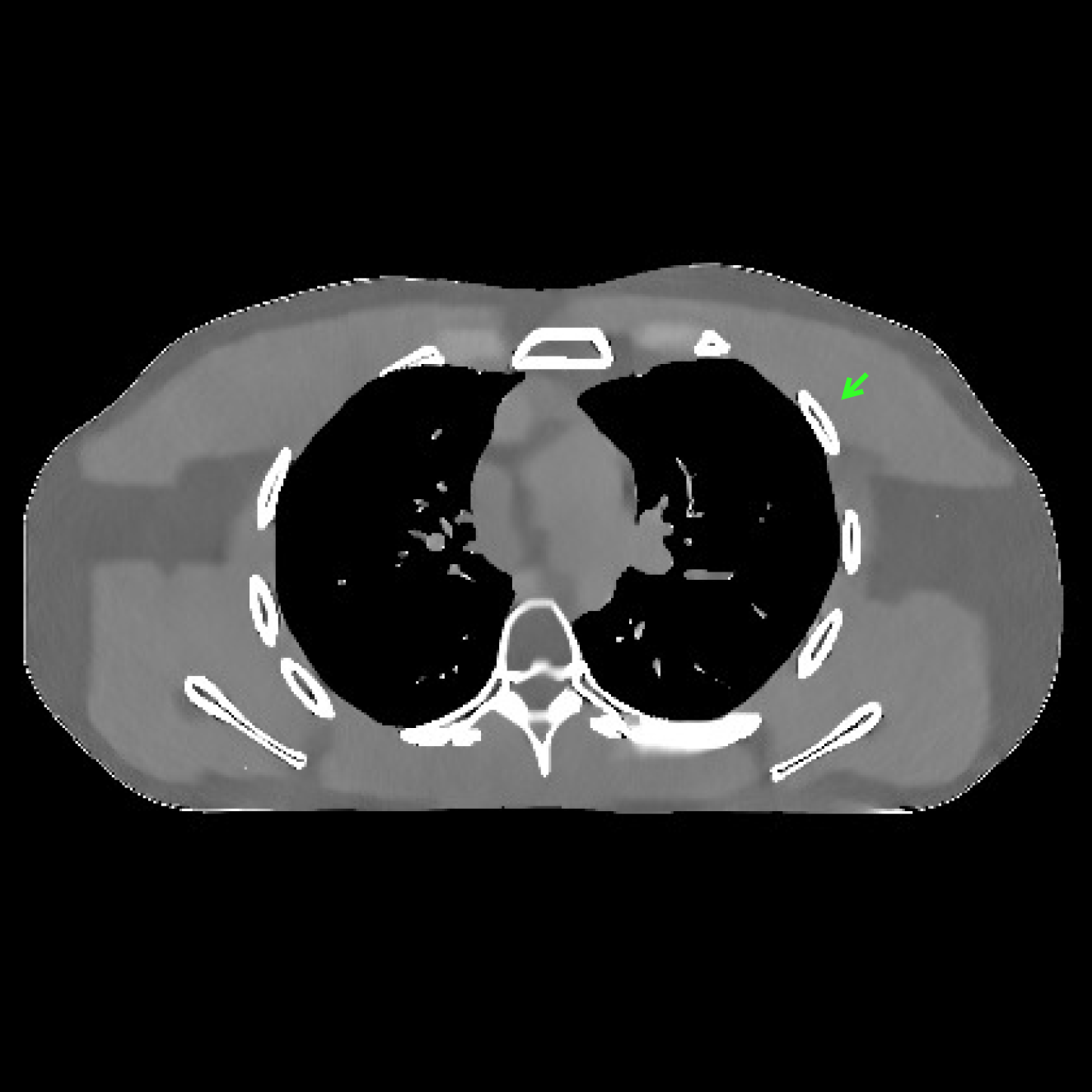}	};
	\spy on (-1.6,-0.2) in node [left] at (-0.9,1.5);		
	\spy on (0,0.25) in node [left] at (1,-1.45);	
	\spy on (1.1,0.5) in node [left] at (1.9,1.5);	
	\end{tikzpicture}
	
	\vspace{-0.1in}	
	\caption{Comparison of 2D reconstructions for PWLS-DL (left) and PWLS-ULTRA ($K=15$, right) at $I_0 = 1\times10^4$.}
	\label{fig:comp_2d_recon}
	\vspace{-0.1in}
\end{figure}

Table~\ref{tab:2d} lists the RMSE and SSIM values for reconstructions with FBP, PWLS-EP, PWLS-DL, PWLS-ST ($\tau_j = 1$), PWLS-ULTRA ($K=15$, $\tau_j = 1$), and PWLS-ULTRA ($K=15$) with the weights $\tau_{j}$.
The adaptive PWLS methods outperform the conventional FBP and the non-adaptive PWLS-EP.
Both PWLS-DL that uses an overcomplete dictionary and PWLS-ULTRA using a union of learned transforms lead to better reconstruction quality than PWLS-ST.
Importantly, PWLS-ULTRA achieves comparable or better image quality than PWLS-DL.
Table~\ref{tab:2d_roi} lists the RMSE values in various ROIs (corresponding to specific tissues) for reconstructions with the six methods. 
The three zoom-ins from left to right in Fig.~\ref{fig:comp_2d_recon} correspond to ROI-1 to ROI-3 in Table ~\ref{tab:2d_roi}, respectively. 
ULTRA achieve lower RMSE in most of these ROIs compared to DL.
Fig.~\ref{fig:comp_2d_recon} compares the reconstructions for PWLS-DL and PWLS-ULTRA without the weights $\tau_{j}$ at $I_0 = 1\times10^4$. 
The ULTRA reconstruction shows fewer artifacts and better clarity of bone and soft tissue edges in the selected ROIs.

%\vspace{-0.1in}
\subsubsection{Runtimes} \label{subsec:2dtime_comp}
To compare the runtimes of various data-driven methods, we ran PWLS-DL, PWLS-ST, and PWLS-ULTRA ($K=15$) (all initialized with the FBP reconstruction) for $200$ outer iterations with $2$ iterations of the image update step and $4$ ordered subsets. For PWLS-ULTRA, we performed the clustering step once every outer iteration.
While the total runtime for the 200 iterations (using a machine with two $2.80$ GHz 10-core Intel Xeon E5-2680 processors) was $95$ minutes for PWLS-DL, it was only $20$ minutes for PWLS-ST and $27$ minutes for PWLS-ULTRA. We observed that PWLS-DL and the proposed methods had similar convergence rates, but the latter were much faster per iteration, thus leading to much lower net runtimes.
The runtime of PWLS-DL was quite equally dominated by the sparse coding (with OMP \cite{pati:93:omp}) and image update steps, whereas for the transform-based methods, the sparse coding and clustering involving simple closed-form solutions and thresholding operations required negligible runtime.
The advantage in runtime was achieved despite using an unoptimized Matlab implementation of PWLS-ST and PWLS-ULTRA, and using an efficient MEX/C implementation for sparse coding  with OMP \cite{pati:93:omp} in PWLS-DL. 
PWLS-DL is far slower for 3D reconstructions with large 3D patches. Hence, we focus our comparisons between the transform learning and dictionary learning-based schemes for 2D LDCT reconstruction.

%\vspace{-0.1in}
\subsection{Low-dose Cone-beam CT Results and Comparisons} \label{sec:resultsd}

We evaluate the performance of various algorithms for reconstructing CT volumes from simulated low-dose cone-beam data. 
Initialized with FDK reconstructions, we ran the PWLS-EP algorithm with edge-preserving parameter $\delta=10$ (HU) and regularization parameter $\beta = 2^{14.5}$ for $50$ iterations with $24$ subsets for both $I_0 = 1 \times 10^4$ and $I_0 = 5 \times 10^3$.
We evaluate PWLS-ST and PWLS-ULTRA without the patch-based weights. 
We also evaluate PWLS-ULTRA with such weights.
Initialized with the PWLS-EP reconstruction, we ran $2$ iterations of the image update step for the proposed methods with $4$ subsets. 
We performed the clustering step once every $20$ outer iterations, which worked well and saved computation.
We chose $\left ( \beta, \gamma, \eta \right )$ for $I_0 = 1 \times 10^4$ and $I_0 = 5 \times 10^3$ as follows:
$\left ( 2.0 \times 10^5, 18, 50 \right )$ and $ \left ( 1.5 \times 10^5, 20, 50 \right )$ for PWLS-ST ($\tau_j = 1$); $\left ( 2.5 \times 10^5, 18, 75 \right )$ and $\left ( 1.5 \times 10^5, 20, 75 \right ) $ for PWLS-ULTRA ($\tau_j = 1$); and $\left ( 1.5 \times 10^4, 18, 75 \right )$ and $\left ( 1.2 \times 10^4, 20,  75\right ) $ for PWLS-ULTRA with the weights $\tau_j$.

Table~\ref{tab:3d} lists the RMSE and SSIM values of the reconstructions with FDK, PWLS-EP, PWLS-ST ($\tau_j = 1$), PWLS-ULTRA ($K=15$, $\tau_j = 1$), and PWLS-ULTRA ($K=15$) with patch-based weights $\tau_j$. Both PWLS-ST and PWLS-ULTRA significantly improve the RMSE and SSIM compared to FDK and the non-adaptive PWLS-EP. 
Importantly, PWLS-ULTRA with a richer union of learned transforms leads to better reconstructions than PWLS-ST with a single learned ST. Incorporating the patch-based weights in PWLS-ULTRA leads to further improvement in reconstruction quality compared to PWLS-ULTRA with uniform weights $\tau_j = 1$ for all patches. In particular, the patch-based weights lead to improved resolution for soft tissues in 3D LDCT reconstructions.

\begin{table}[!t]	
	\centering
	\caption{RMSE (HU) and SSIM of 3D (cone-beam) reconstructions with FDK, PWLS-EP, PWLS-ST, PWLS-ULTRA ($K=15$), and PWLS-ULTRA ($K=15$) with patch-based weights ($\tau_{j}$), for two incident photon intensities}
	\label{tab:3d}	
	\vspace{-0.1in}
	\begin{tabular}{cccccc}			
		\toprule
		Intensity                                      &    FDK          &  EP        &  ST       & ULTRA    & ULTRA-$\{\tau_j\}$    \\
		\midrule
		\multirow{2}{*}  {$1\times10^4$}    &67.8       & 34.6      & 32.1       & 30.7        & \bf{29.2}    \\
		\cmidrule{2-6}                                &0.536     & 0.940      &0.976     & 0.978     & \bf{0.981}    \\
		\midrule
		\multirow{2}{*}  {$5\times10^3$}    &89.0       & 41.1       & 37.3       & 35.7     & \bf{34.2}    \\
		\cmidrule{2-6}                                &0.463     & 0.921     &0.967      &  0.970   & \bf{0.974}    \\
		\bottomrule
	\end{tabular}
	\vspace{-0.1in}
\end{table}

\begin{figure}[!t]
	\centering  	
	\begin{tikzpicture}
	[spy using outlines={rectangle,green,magnification=2,size=10mm, connect spies}]
	\node {\includegraphics[width=0.21\textwidth]{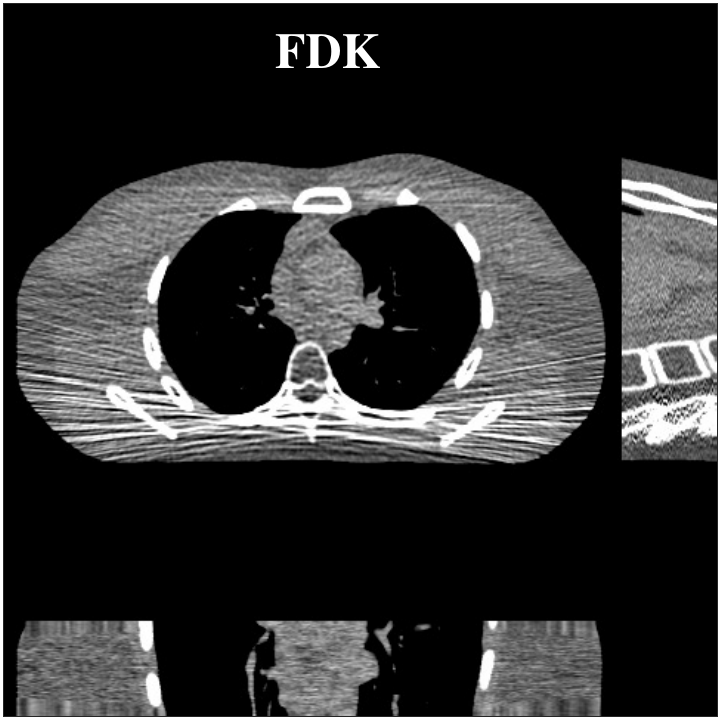}	};
	\spy on (0.87,0.12) in node [left] at (2.1,-1.2);	
	\spy on (-0.25,0.4) in node [left] at (2.1,1.65);			
	\end{tikzpicture}
	\includegraphics[width=0.25\textwidth]{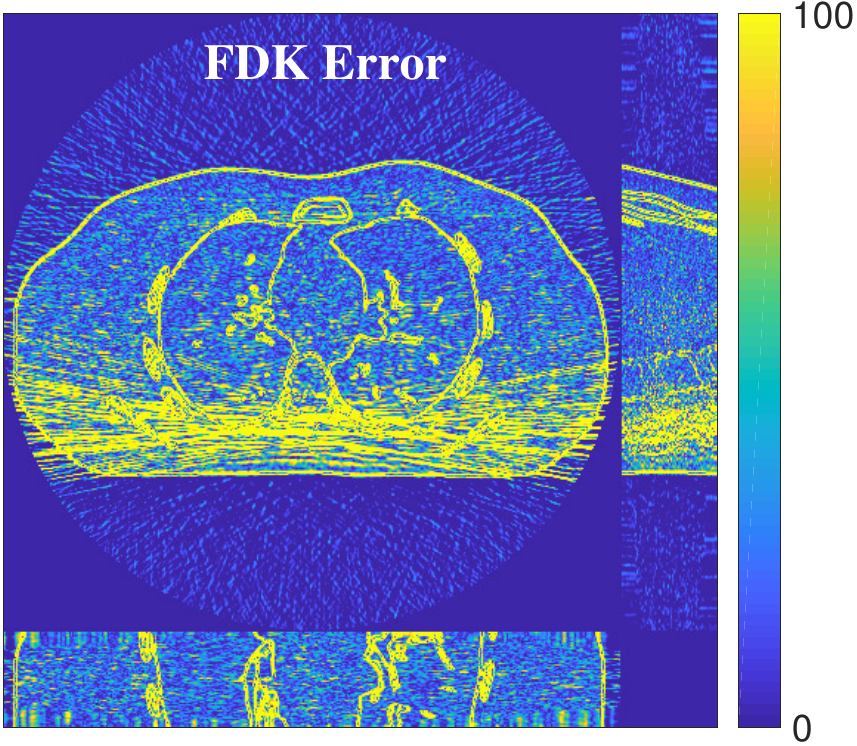}		
	\begin{tikzpicture}
	[spy using outlines={rectangle,green,magnification=2,size=10mm, connect spies}]
	\node {\includegraphics[width=0.21\textwidth]{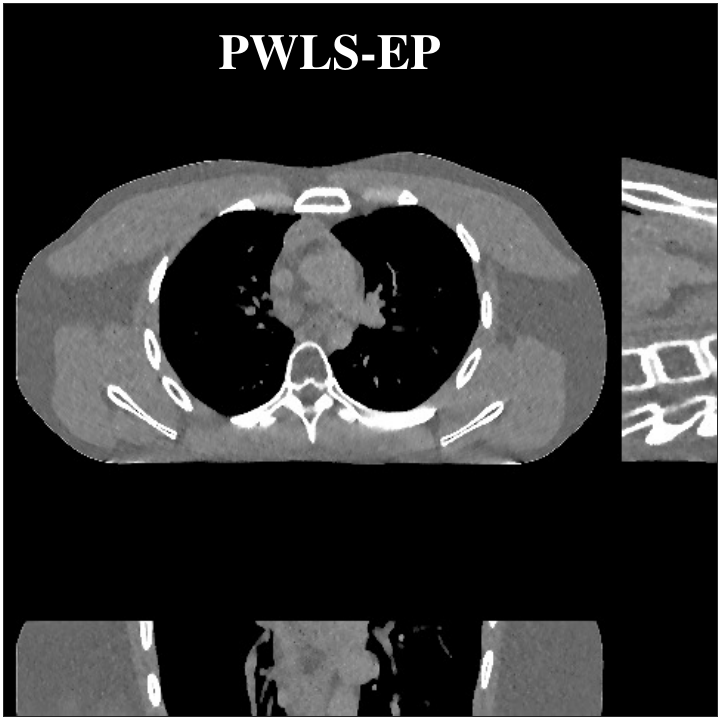}	};
	\spy on (0.87,0.12) in node [left] at (2.1,-1.2);	
	\spy on (-0.25,0.4) in node [left] at (2.1,1.65);			
	\end{tikzpicture}
	\includegraphics[width=0.25\textwidth]{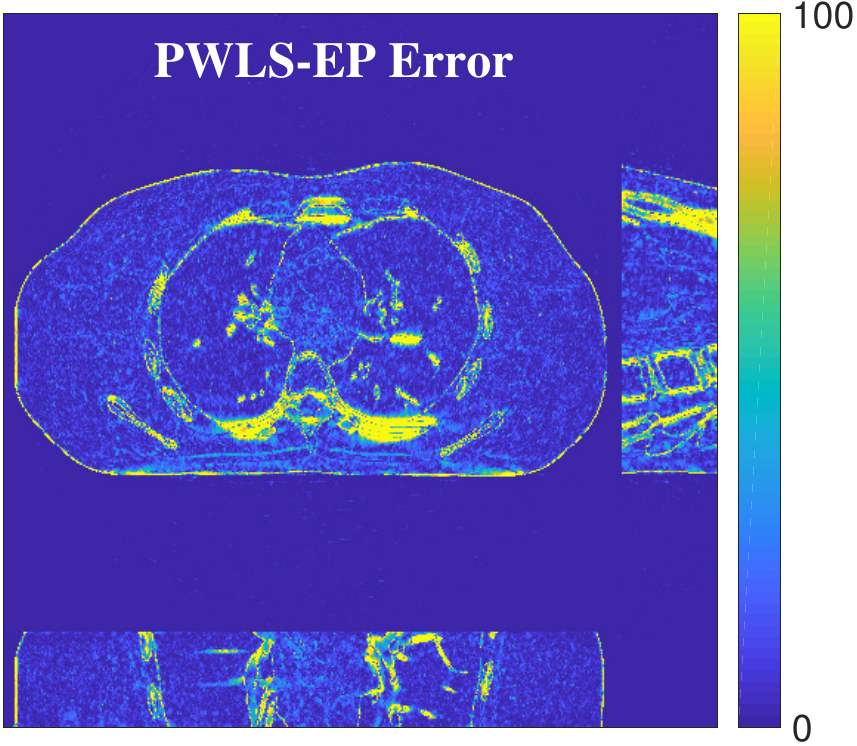}
	\begin{tikzpicture}
	[spy using outlines={rectangle,green,magnification=2,size=10mm, connect spies}]
	\node {\includegraphics[width=0.21\textwidth]{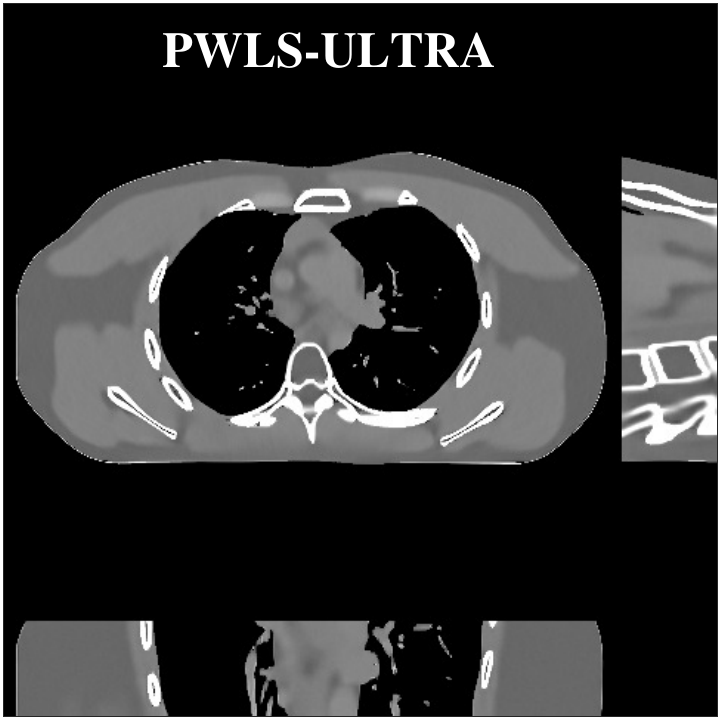}	};
	\spy on (0.87,0.12) in node [left] at (2.1,-1.2);	
	\spy on (-0.25,0.4) in node [left] at (2.1,1.65);			
	\end{tikzpicture}
	\includegraphics[width=0.25\textwidth]{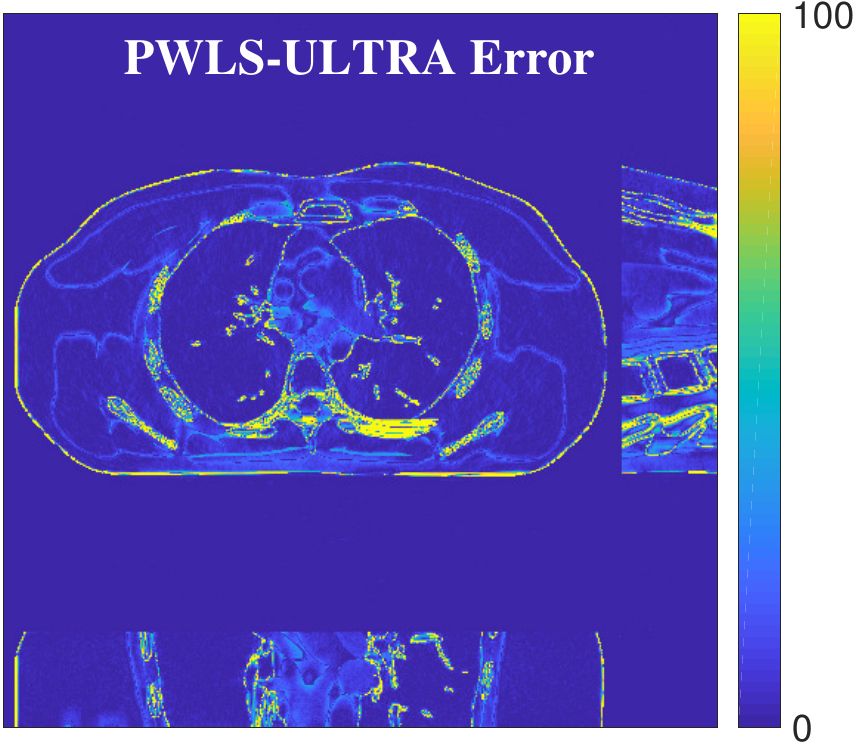}	
	\vspace{-0.3in}	
	\caption{Comparison of the reconstructions and corresponding error images (shown for the central axial, sagittal, and coronal planes) for FDK, PWLS-EP, and PWLS-ULTRA ($K=15$) with patch-based weights at $I_0 = 1 \times 10^4$. The unit of the display window of the error images is HU.}
	\label{fig:comp}
	\vspace{-0.05in}
\end{figure}

\begin{figure}[!h]
	\centering  	
	\includegraphics[width=0.49\textwidth]{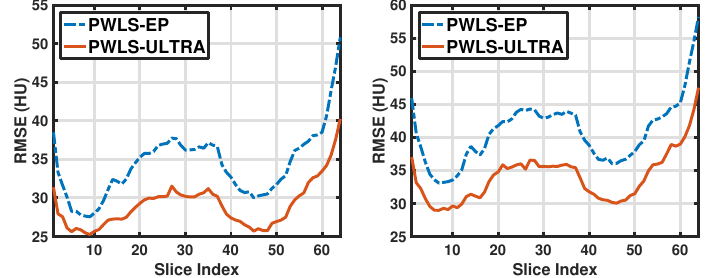}			
	\vspace{-0.25in}
	\caption{RMSE of each axial slice in the PWLS-EP and PWLS-ULTRA reconstructions for $I_0 = 1\times10^4$ (left) and $I_0 = 5\times10^3$ (right).}
	\label{fig:rmseprofile}
	\vspace{-0.2in}
\end{figure}

Fig.~\ref{fig:comp} shows the reconstructions and the corresponding error (magnitudes) images (shown for the central axial, sagittal, and coronal planes) for FDK, PWLS-EP, and PWLS-ULTRA ($K=15$) with the patch-based weights. Compared to FDK and PWLS-EP, PWLS-ULTRA significantly improves image quality by reducing noise and preserving structural details (see zoom-ins).
Fig.~\ref{fig:rmseprofile} shows the RMSE for each axial slice in the PWLS-EP and PWLS-ULTRA (with the weights $\tau_j$) reconstructions. PWLS-ULTRA clearly provides large improvements in RMSE for many slices, with greater improvements near the central slice.

%\vspace{-0.1in}

\subsection{Results for Clinical Data: Chest and Abdomen Scans} \label{sec:resultse}

We reconstructed the chest volume from helical CT data. For PWLS-EP, we used the same parameter settings as used for this data in prior work \cite{nien:16:rla}. 
Initializing with the PWLS-EP reconstruction, we ran the PWLS-ULTRA ($K=5$) method with the weights $\tau_j$ for $78$ outer iterations with $3$ iterations of the image update step and $4$ subsets. We performed clustering once every $10$ outer iterations. We chose $\beta = 2 \times 10^5$ and $\gamma = 25$ for PWLS-ULTRA  to obtain good visual quality of the reconstruction.
We used the transforms learned from the XCAT phantom volume with $\eta = 100$ to obtain reconstructions with PWLS-ULTRA for the clinical chest CT data. The supplement shows that transforms learned from the XCAT phantom provide similar visual reconstructions as transforms learned from the PWLS-EP reconstruction of the chest data. This suggests that the transform learning algorithm may extract quite general and effective image features without requiring a very closely matched training dataset, which is a key distinction from the PICCS and ndiNLM-type methods \cite{chen:08:pic, ramirezgiraldo:11:npi, chen:12:tri, lauzier13:cos, ma:11:ldc, zhang:17:aon}.

\begin{figure*}[!t]
	\centering
	\includegraphics[width=1\textwidth]{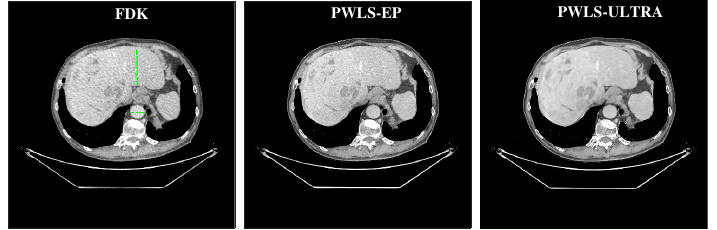}			
	\vspace{-0.2in}
	\caption{Chest reconstructions (shown for central axial plane) from helical CT data, with the FDK, PWLS-EP, and PWLS-ULTRA ($K=5$) methods.}
	\label{fig:chest}
	\vspace{-0.15in}
\end{figure*}

\begin{figure*}[!h]
	\centering  
	\begin{tikzpicture}
	[spy using outlines={rectangle,green,magnification=2,size=10mm, connect spies}]
	\node {	\subfigure[1X]{\includegraphics[width=0.22\textwidth]{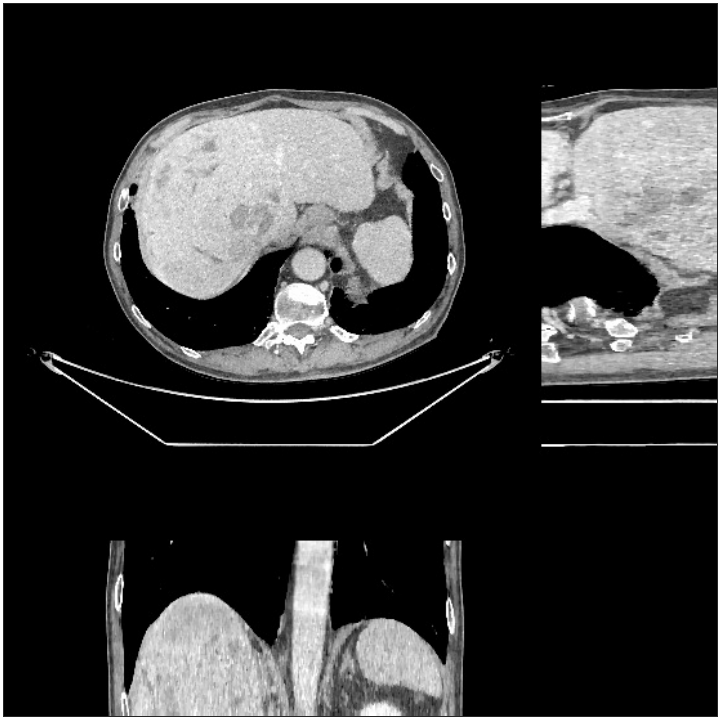}	}};
	\spy on (1.4,0.6) in node [left] at (1.8,-1.0);	
	\end{tikzpicture}
	\begin{tikzpicture}
	[spy using outlines={rectangle,green,magnification=2,size=10mm, connect spies}]
	\node {	\subfigure[2X]{\includegraphics[width=0.22\textwidth]{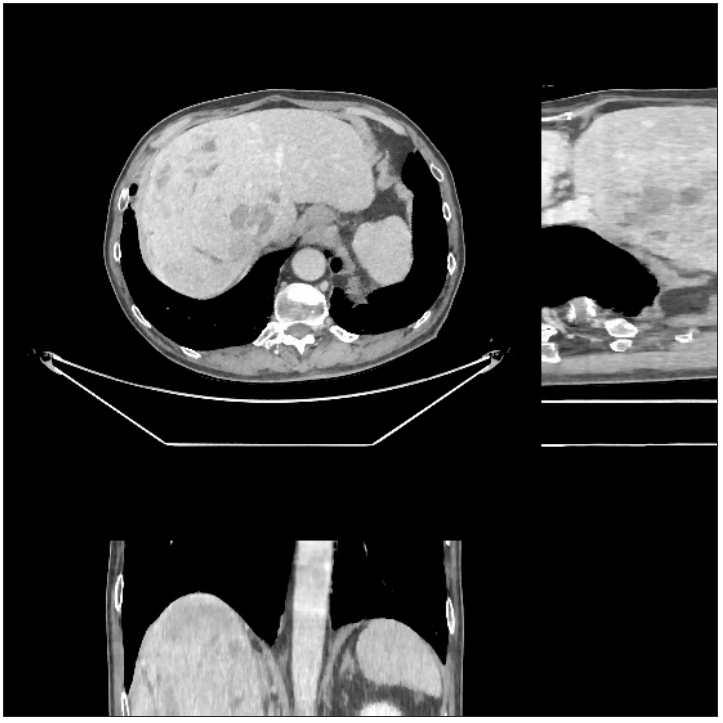}	}};
	\spy on (1.4,0.5) in node [left] at (1.8,-1.1);	
	\end{tikzpicture}	
	\begin{tikzpicture}
	[spy using outlines={rectangle,green,magnification=2,size=10mm, connect spies}]
	\node {	\subfigure[0.5X]{\includegraphics[width=0.22\textwidth]{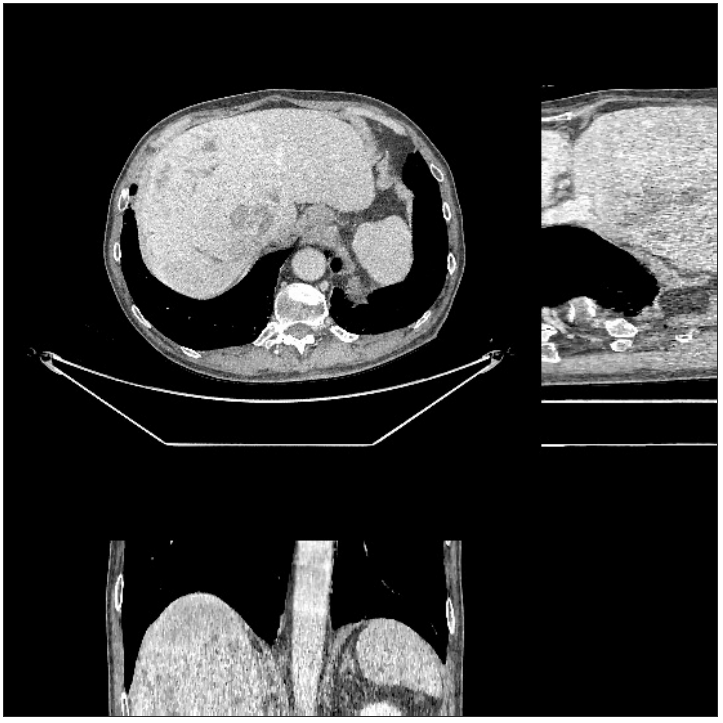}	}};
	\spy on (1.4,0.5) in node [left] at (1.8,-1.1);	
	\end{tikzpicture}
	\begin{tikzpicture}
	[spy using outlines={rectangle,green,magnification=2,size=10mm, connect spies}]
	\node {	\subfigure[0.25X]{\includegraphics[width=0.22\textwidth]{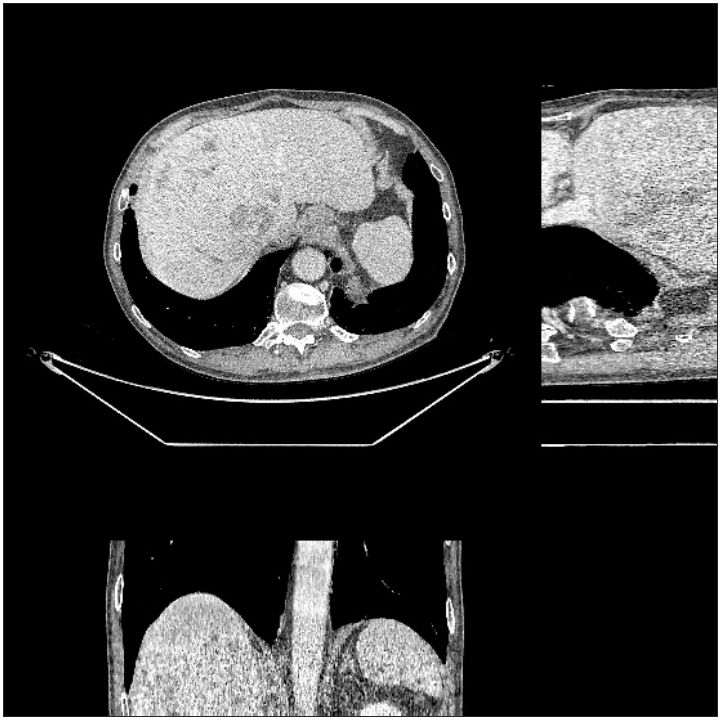}	}};
	\spy on (1.4,0.5) in node [left] at (1.8,-1.1);	
	\end{tikzpicture}
	\vspace{-0.2in}
	\caption{Chest reconstructions (shown for the central axial, sagittal, and coronal planes in the 3D volume) for PWLS-EP with different regularization strengths. $1$X denotes the chosen regularization parameter in \cite{nien:16:rla} that  provides a good trade-off between image resolution and noise reduction. The $2$X, $0.5$X, and $0.25$X denote scaling of the parameter $\beta$ over the $1$X case. }
	\label{fig:chest:ep_paras}
	\vspace{-0.15in}
\end{figure*}

\begin{figure*}[!h]
	\centering  	
	\begin{tikzpicture}
	[spy using outlines={rectangle,green,magnification=2,size=10mm, connect spies}]
	\node {	\subfigure[$\beta = 2 \times 10^5, \gamma = 25$]{\includegraphics[width=0.22\textwidth]{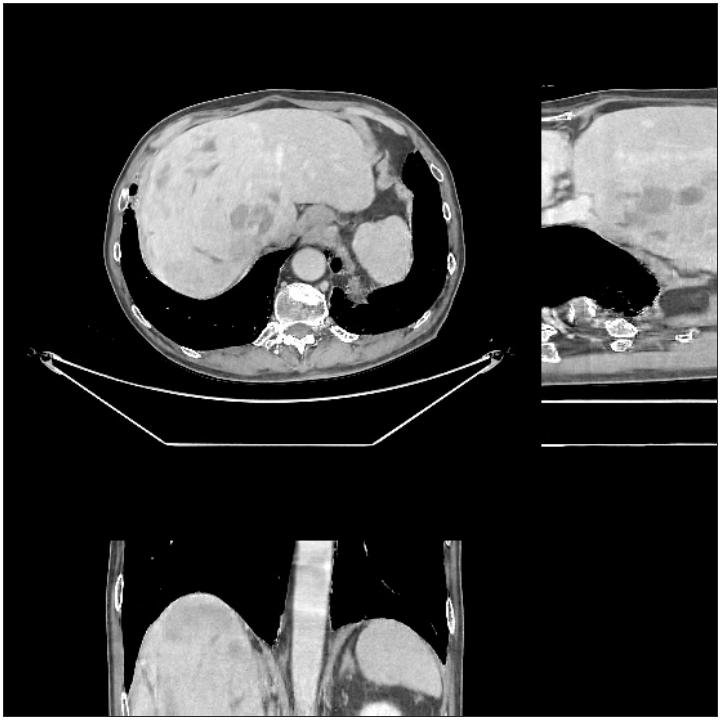}		}};
	\spy on (1.4,0.6) in node [left] at (1.8,-1.0);	
	\end{tikzpicture}
	\begin{tikzpicture}
	[spy using outlines={rectangle,green,magnification=2,size=10mm, connect spies}]
	\node {	\subfigure[$\beta = 2 \times 10^5, \gamma = 20$]{\includegraphics[width=0.22\textwidth]{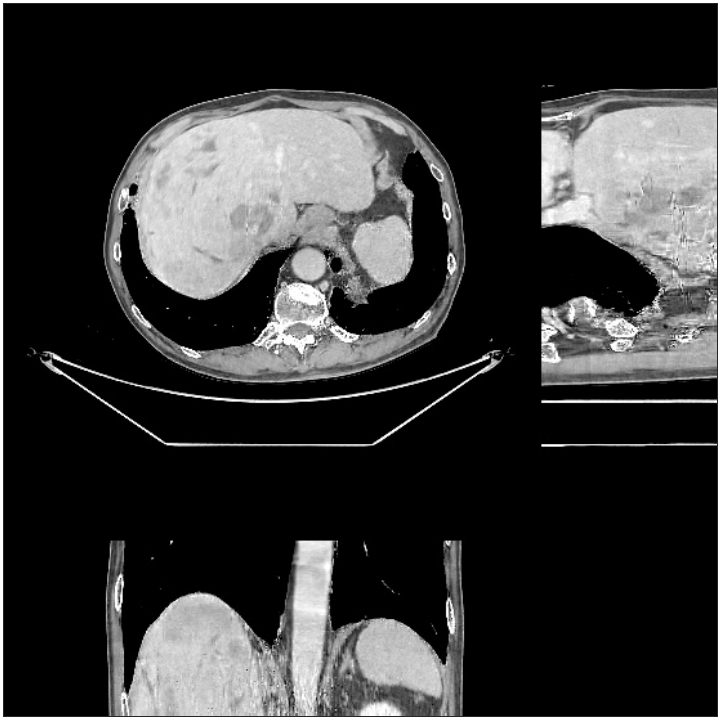}		}};
	\spy on (1.4,0.5) in node [left] at (1.8,-1.1);	
	\end{tikzpicture}
	\begin{tikzpicture}
	[spy using outlines={rectangle,green,magnification=2,size=10mm, connect spies}]
	\node {	\subfigure[$\beta = 3 \times 10^5, \gamma = 20$]{\includegraphics[width=0.22\textwidth]{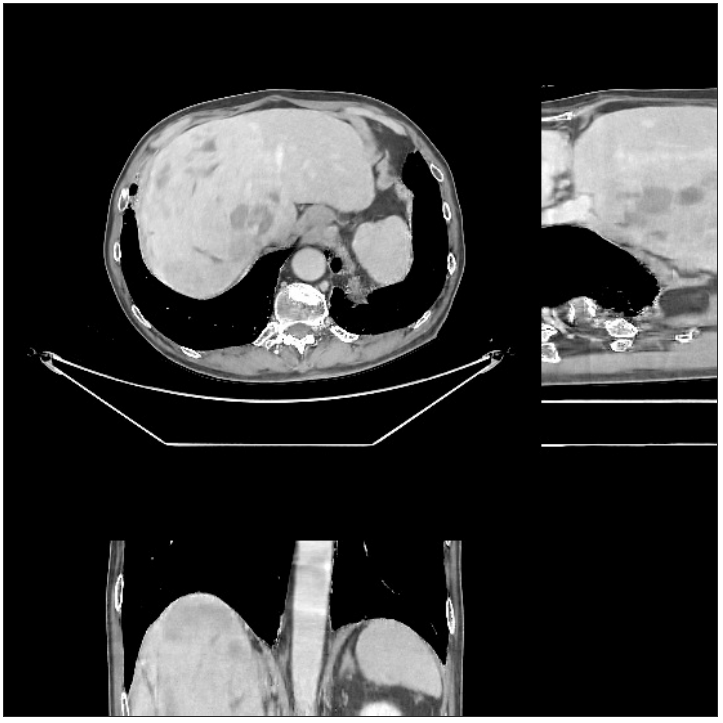}		}};
	\spy on (1.4,0.5) in node [left] at (1.8,-1.1);	
	\end{tikzpicture}
	\begin{tikzpicture}
	[spy using outlines={rectangle,green,magnification=2,size=10mm, connect spies}]
	\node {	\subfigure[$\beta = 3 \times 10^5, \gamma = 25$]{\includegraphics[width=0.22\textwidth]{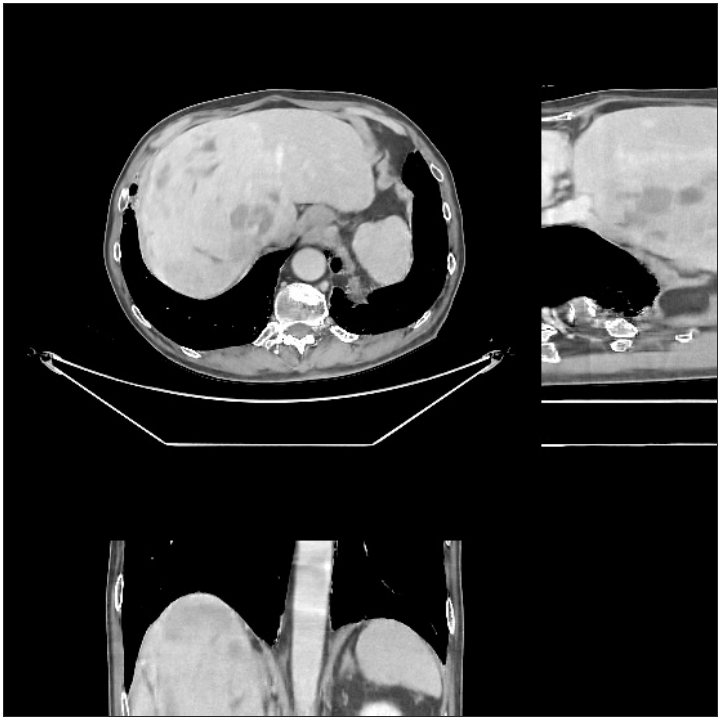}		}};
	\spy on (1.4,0.5) in node [left] at (1.8,-1.1);	
	\end{tikzpicture}
	\vspace{-0.2in}
	\caption{Chest reconstructions (shown for the central axial, sagittal, and coronal planes in the 3D volume) for PWLS-ULTRA ($K=5$) with different parameter combinations. Larger regularization strength $\beta$ would achieve more noise reduction but simultaneously lower spatial resolution, e.g., compare (a) and (d); larger values of $\gamma$ would achieve lower sparsities and more noise reduction but potentially oversmooth the image, e.g., compare (c) and (d). }
	\label{fig:chest:ultra_paras}
	\vspace{-0.15in}
\end{figure*}

\begin{figure*}[!h]
	\centering  	
	\begin{tikzpicture}
	[spy using outlines={rectangle,green,magnification=2,size=10mm, connect spies}]
	\node {	\subfigure[$150$ mA, PWLS-EP]{\includegraphics[width=0.22\textwidth]{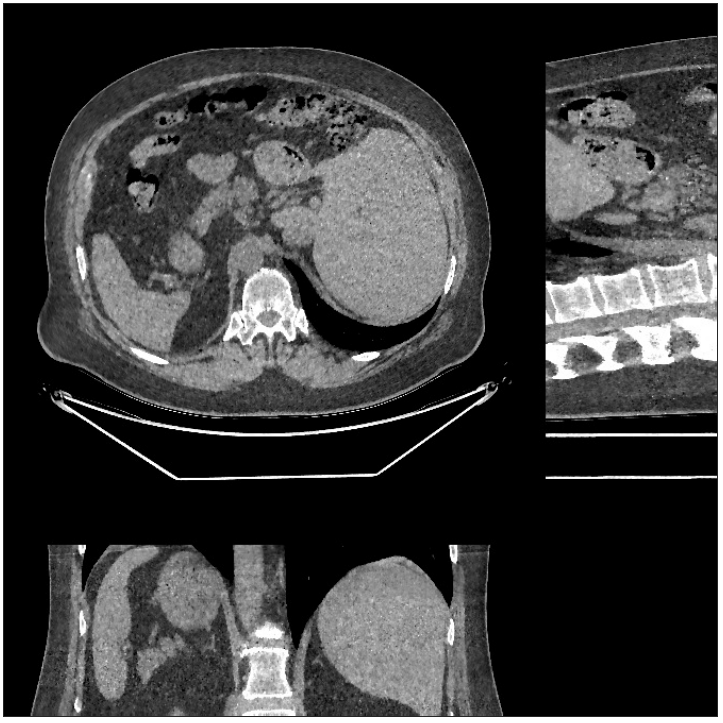}		}};
	%\spy on (-0.6,1.2) in node [left] at (1.8,-1.0);	
	\spy on (-0.6,0.6) in node [left] at (1.8,-1.0);	
	\end{tikzpicture}
	\begin{tikzpicture}
	[spy using outlines={rectangle,green,magnification=2,size=10mm, connect spies}]
	\node {	\subfigure[$150$ mA, PWLS-ULTRA-$\{\tau_j\}$]{\includegraphics[width=0.22\textwidth]{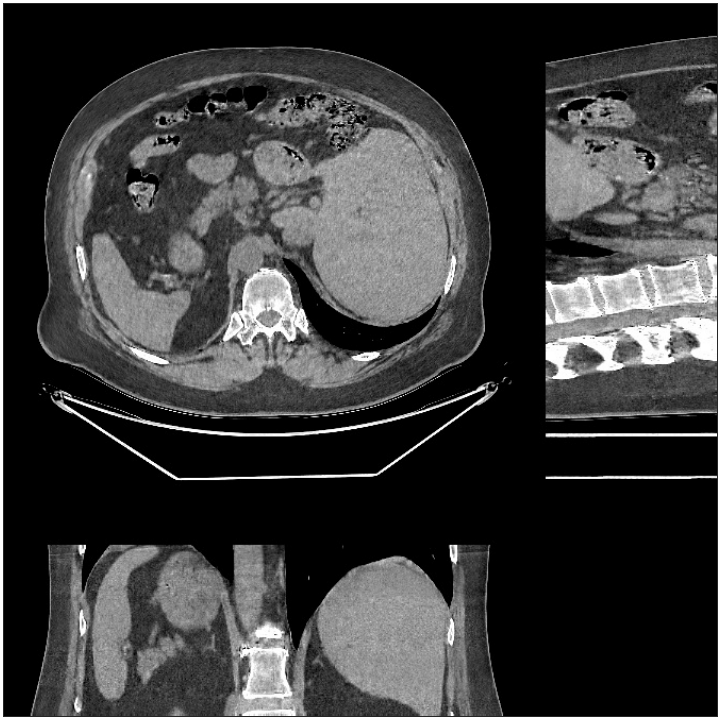}		}};
	%\spy on (-0.6,1.1) in node [left] at (1.8,-1.1);	
	\spy on (-0.6,0.5) in node [left] at (1.8,-1.1);	
	\end{tikzpicture}
	\begin{tikzpicture}
	[spy using outlines={rectangle,green,magnification=2,size=10mm, connect spies}]
	\node {	\subfigure[$35$ mA, PWLS-EP]{\includegraphics[width=0.22\textwidth]{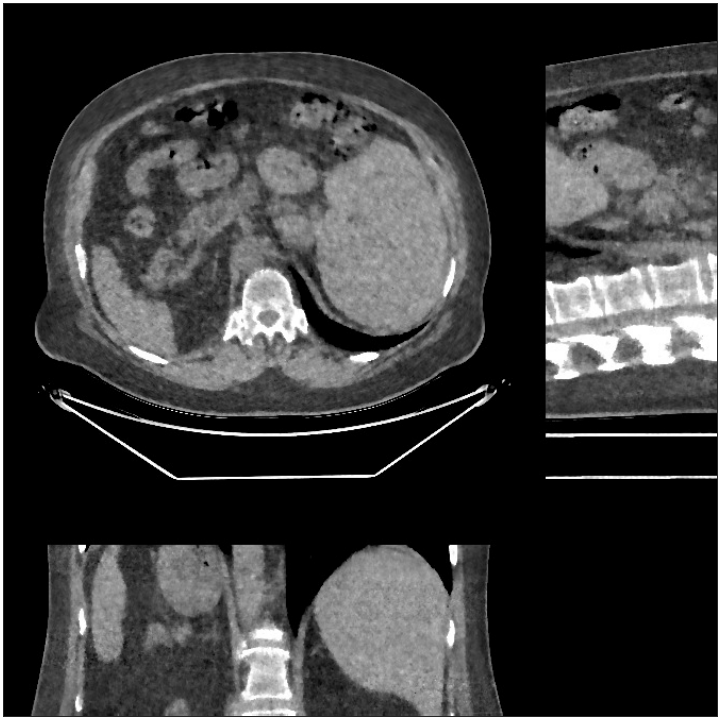}		}};
	\spy on (-0.6,0.5) in node [left] at (1.8,-1.1);	
	\end{tikzpicture}
	\begin{tikzpicture}
	[spy using outlines={rectangle,green,magnification=2,size=10mm, connect spies}]
	\node {	\subfigure[$35$ mA, PWLS-ULTRA-$\{\tau_j\}$]{\includegraphics[width=0.22\textwidth]{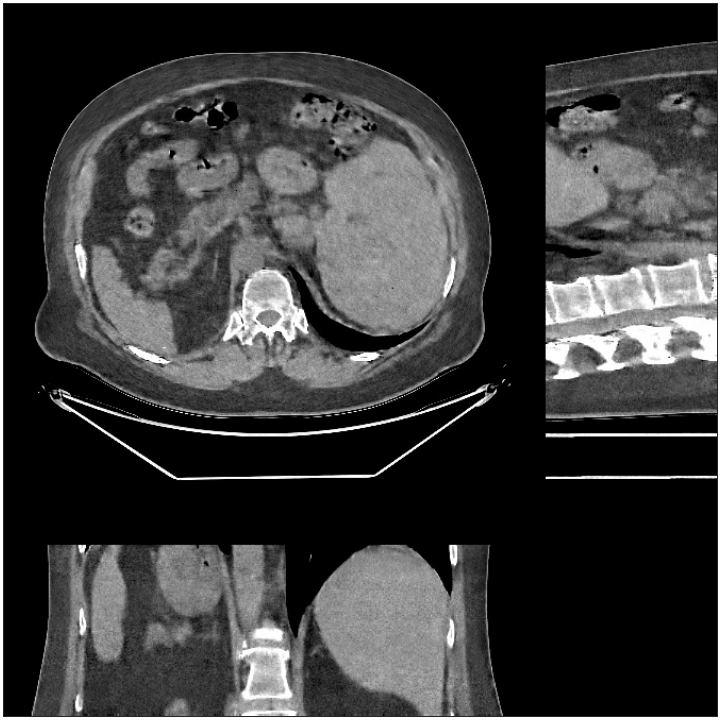}		}};
	\spy on (-0.6,0.5) in node [left] at (1.8,-1.1);	
	\end{tikzpicture}
	\vspace{-0.2in}
	\caption{Abdomen reconstructions (shown for the central axial, sagittal, and coronal planes, and air cropped) from low-dose ($120$kVp, $150$mA and $35$mA with rotation time $0.8$ seconds) helical CT data (the same patient) for PWLS-EP and PWLS-ULTRA with patch-based weights ($K=5$).}
	\label{fig:abdomen}
	\vspace{-0.15in}
\end{figure*}

Fig.~\ref{fig:chest} shows the reconstructions (shown for the central axial plane in the 3D volume) for FDK (provided by GE Healthcare), PWLS-EP (corresponds to Fig.~\ref{fig:chest:ep_paras}(a)), and PWLS-ULTRA with $K=5$ (corresponds to Fig.~\ref{fig:chest:ultra_paras}(a)).
The PWLS-ULTRA reconstruction has lower artifacts and noise. Moreover, the image features and edges are better reconstructed by PWLS-ULTRA than by PWLS-EP or FDK.

\begin{figure}[!t]
	\centering  	
	\includegraphics[width=0.49\textwidth]{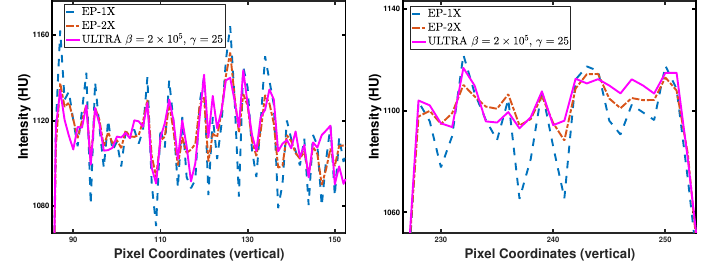}	
	\vspace{-0.25in}
	\caption{Vertical (left) and horizontal (right) profiles of chest reconstructions (plotted from the central axial slice) for the PWLS-EP and PWLS-ULTRA methods. The profile locations are shown in green lines in  Fig.~\ref{fig:chest}. }
	\label{fig:chest:profile}
	\vspace{-0.2in}
\end{figure}

Fig.~\ref{fig:chest:ep_paras} shows the reconstructions (shown for the central axial, sagittal, and coronal planes in the 3D volume) for PWLS-EP with different regularization strengths $\beta$, denoted as a multiplicative factor of the parameter value in Fig.~\ref{fig:chest}. 
Fig.~\ref{fig:chest:ultra_paras} shows the reconstructions for PWLS-ULTRA (with patch-based weights) with different parameter combinations. 
For the sagittal and coronal planes, we show the central $135$ out of $222$ axial slices. 
Larger regularization strengths $\beta$ would achieve more noise reduction but simultaneously lower spatial resolution in PWLS-EP and PWLS-ULTRA, e.g., compare Fig.~\ref{fig:chest:ep_paras} and Figs.~\ref{fig:chest:ultra_paras}(a) and (d). 
Larger values of $\gamma$ would achieve lower sparsities and more noise reduction but potentially oversmooth the image, e.g., compare Figs.~\ref{fig:chest:ultra_paras}(c) and (d). 
Small values of $\gamma$ may introduce additional spurious noise in the PWLS-ULTRA reconstruction (compare Figs.~\ref{fig:chest:ultra_paras}(a) and (b)).
Fig.~\ref{fig:chest:profile} shows profiles of chest reconstructions (plotted from the central axial slice) for the PWLS-EP and PWLS-ULTRA methods. The profile locations are shown in green lines in Fig.~\ref{fig:chest}.
Both PWLS-EP with regularization strength $2$X and PWLS-ULTRA (with patch-based weights) in Fig.~\ref{fig:chest:ultra_paras}(a) have lower noise than the PWLS-EP with regularization strength $1$X. 
Though the spatial resolution of PWLS-EP with regularization strength $2$X is close to PWLS-ULTRA in the selected soft-tissue regions, PWLS-ULTRA reconstructs bone and spine areas with higher resolution, and preserves small features better (compare the zoomed-in areas in Fig.~\ref{fig:chest:ep_paras} and Fig.~\ref{fig:chest:ultra_paras}).

We reconstructed the abdomen volume from low-dose helical CT data. With an initialization of zeros, we ran the PWLS-EP algorithm with $\beta = 2^{18.0}$ and $\beta = 2^{19.0}$ for $20$ iterations with $12$ subsets for the $150$ mA and $35$ mA scans, respectively. For PWLS-ULTRA, we chose $\beta =  1 \times 10^5, \gamma = 25$ for the $150$ mA scan, $\beta =  1.5 \times 10^5, \gamma = 30$ for the $35$ mA scan, and ran it for $50$ outer iterations. The other parameter settings and the transform were the same as those used for the chest scan.

Fig.~\ref{fig:abdomen} shows the reconstructions (shown for the central axial, sagittal, and coronal planes in the 3D volume) for PWLS-EP and PWLS-ULTRA with patch-based weights ($K=5$) from low-dose abdomen scans. For the sagittal and coronal planes, we show the central $160$ out of $200$ axial slices. The supplement provides PWLS-EP reconstructions with different regularization strengths. 
The PWLS-ULTRA reconstructions in Fig.~\ref{fig:abdomen} have reduced noise as well as higher resolution, better structural details and shaper image edges than the PWLS-EP results.
These results are further example of the potential performance of the proposed PWLS-ULTRA method in clinical settings.

%\vspace{-0.1in}
\subsection{Comparison to Oracle Clustering Scheme} \label{sec:resultsf}

We consider the 3D cone-beam CT data in Section \ref{sec:resultsd} with $I_0 = 1 \times 10^4$, and compare the PWLS-ULTRA ($K=15$) method without patch-based weights to an oracle PWLS-ULTRA scheme without patch-based weights, where the cluster memberships are pre-determined (and fixed during reconstruction) by performing the sparse coding and clustering step (with the learned transforms) on the patches of the reference or ground truth volume.
The oracle scheme thus uses the best possible estimate of the cluster memberships. 
Otherwise, we used the same parameters for the two cases.
Fig.~\ref{fig:true_cluster} compares the reconstructions for the two cases. The proposed PWLS-ULTRA underperforms the oracle scheme by only $1.7$ HU. The more precise clustering leads to sharper edges for the latter method. This also suggests that there is room for potentially improving the proposed clustering-based PWLS-ULTRA scheme, which could be pursued in future works.

\begin{figure}[!h]
	\vspace{-0.1in}
	\centering  	
	\begin{tikzpicture}
	[spy using outlines={rectangle,green,magnification=2,size=9mm, connect spies}]
	\node {\includegraphics[width=0.225\textwidth]{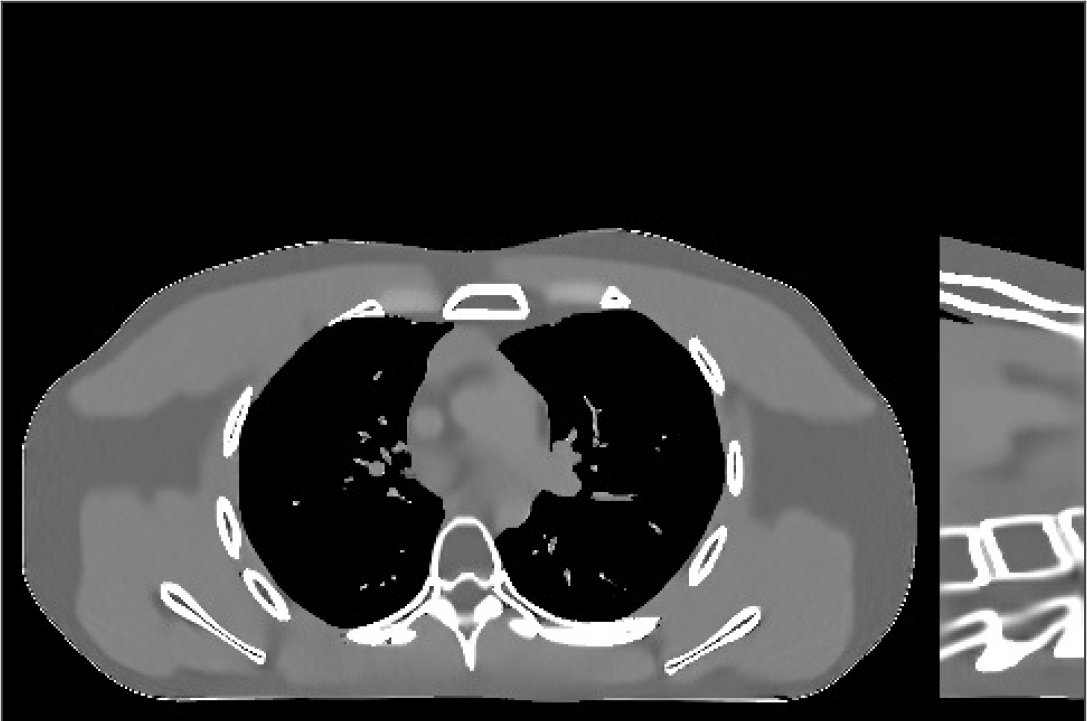}	};
	\spy on (-0.25,-0.9) in node [left] at (-1,0.87);	
	\spy on (1.75,-0.8) in node [left] at (1.5,0.87);		
	\end{tikzpicture}
	\begin{tikzpicture}
	[spy using outlines={rectangle,green,magnification=2,size=9mm, connect spies}]
	\node {\includegraphics[width=0.225\textwidth]{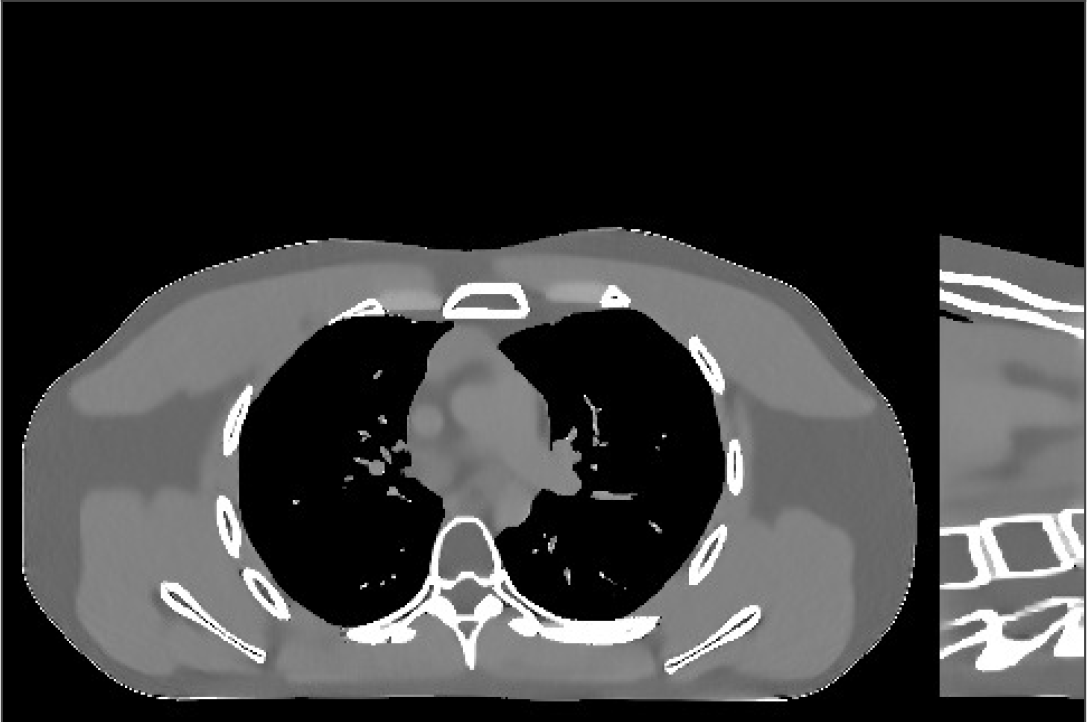}	};
	\spy on (-0.25,-0.9) in node [left] at (-1,0.87);	
	\spy on (1.75,-0.8) in node [left] at (1.5,0.87);		
	\end{tikzpicture}
	\vspace{-0.1in}
	\caption{Reconstruction with PWLS-ULTRA ($K=15$) without weights $\tau_j$ (left) at $I_0 = 1\times10^4$ compared to the reconstruction with the oracle scheme without weights $\tau_j$ (right), where the cluster memberships were pre-determined from the ground truth. RMSE and SSIM values of $30.7$ and $0.978$ (left), and $29.0$ and $0.982$ (right) respectively, for the volumes, indicates that more precise clustering can provide better reconstructions and sharper edges (see zoom-ins).}
	\label{fig:true_cluster} 
	\vspace{-0.2in}
\end{figure}

%\vspace{-0.15in}

\section{Conclusions}
\label{sec:conclusions}

We presented the PWLS-ST and PWLS-ULTRA methods for low-dose CT imaging, combining conventional penalized weighted least squares reconstruction with regularization based on pre-learned sparsifying transforms. 
Experimental results with 2D and 3D axial CT scans of the XCAT phantom and 3D helical chest and abdomen scans show that for both normal-dose and low-dose levels, the proposed  methods provide high quality image reconstructions compared to conventional techniques such as FBP or PWLS reconstruction with a nonadaptive edge-preserving regularizer. 
The ULTRA scheme with a richer union of transforms model provides better reconstruction of various features such as bones, specific soft tissues, and edges, compared to the proposed PWLS-ST.
Finally, the proposed approach achieves comparable or better image quality compared to learned overcomplete synthesis dictionaries, but importantly, is much faster (computationally more efficient).
We leave the investigation of convergence guarantees and automating the parameter selection for the proposed PWLS algorithms to future work. 
The field of transform learning is rapidly growing, and we hope to investigate new transform learning-based LDCT reconstruction methods, such as involving rotationally invariant transforms \cite{wen:17:fla}, or online transform learning \cite{ravishankar:15:ost-1, ravishankar:15:ost-2}, etc., in future work.

%\vspace{-0.1in}
\section{Acknowledgments}

The authors thank GE Healthcare for supplying the helical chest and abdomen data used in this work. 
The authors also thank Dr. Hung Nien for his feedback.

%\vspace{-0.15in}
\bibliographystyle{IEEEtran}
\bibliography{refs}	
% -------------------------------------------------------------------------

% -------------------------------------------------------------------------
\newpage
\clearpage

%%%%%%%%%%%%%%%%%%%%%%%%%%%%%%%%%%%%%%%%%%%%%%%%%%
% BEGIN SUPPLEMENT
%%%%%%%%%%%%%%%%%%%%%%%%%%%%%%%%%%%%%%%%%%%%%%%%%%

% Title
{
	\twocolumn[
	\begin{center}
		\Huge PWLS-ULTRA: An Efficient Clustering and Learning-Based Approach for Low-Dose 3D CT Image Reconstruction: Supplementary Material
		\vspace{0.4in}
	\end{center}]
}

	This supplement provides additional details and experimental results to accompany our manuscript \cite{zheng:17:pua}.

	\setcounter{section}{6}
	\section{Computing $\D_\R$ in Algorithm $1$}
	
	Recall the following definition of $\D_\R$ in (11):
		  \begin{equation}
		  	\setcounter{equation}{14}
	\D_\R  \triangleq 2 \beta \bigg\{\max_k \lambda_{\max}(\omg_k^{T} \omg_k) \bigg\} \sum_{k=1}^{K}  \sum_{j\in C_k}\tau_j \P_j^{T}\P_j.
	\end{equation}
When $\{\tau_j\}$ are all identical (or by replacing them with $\max_j \tau_j$ above for a looser majorizer), $\sum_{k=1}^{K} \sum_{j\in C_k}\P_j^{T}\P_j   = \sum_{j=1}^{\tilde{N}}\P_j^{T}\P_j  \in \mathbb{R}^{N_p \times N_p }$ is a diagonal matrix with the diagonal entries corresponding to image voxel locations and their values being the total number of image patches overlapping each voxel. Moreover, if the patches are periodically positioned with a stride of 1 voxel along each dimension and wrap around at image boundaries, then $\sum_{j=1}^{\tilde{N}}\P_j^{T}\P_j = l \I $. In this case, $\D_\R  = 2 \beta \{\max_k \lambda_{\max}(\omg_k^{T} \omg_k) \} \{\max_j \tau_j \}l \I$. More generally, when $\tau_j$ values differ, we compute $\sum_{k=1}^{K}  \sum_{j\in C_k}\tau_j \P_j^{T}\P_j$ voxel-wise by summing the $\tau_j$ values of the patches overlapping the voxel.

\setcounter{section}{7}
	\section{Additional Experimental Results}

\begin{figure}[!t]
	\setcounter{figure}{12}
	\centering  	
	\begin{tikzpicture}
	[spy using outlines={rectangle,green,magnification=2,size=10mm, connect spies}]
	\node {\includegraphics[width=0.21\textwidth]{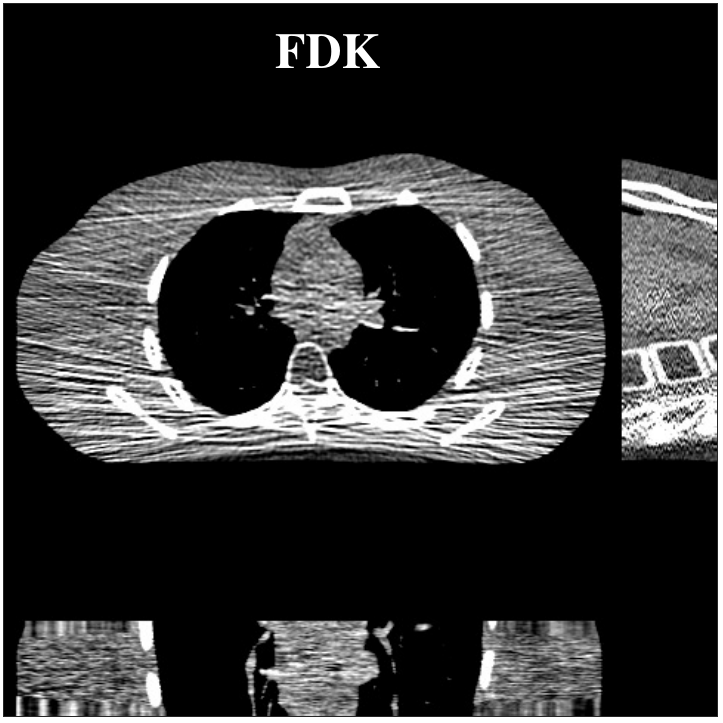}	};
	\spy on (0.87,0.12) in node [left] at (2.1,-1.2);	
	\spy on (-0.25,0.4) in node [left] at (2.1,1.65);		
	\end{tikzpicture}
	\includegraphics[width=0.25\textwidth]{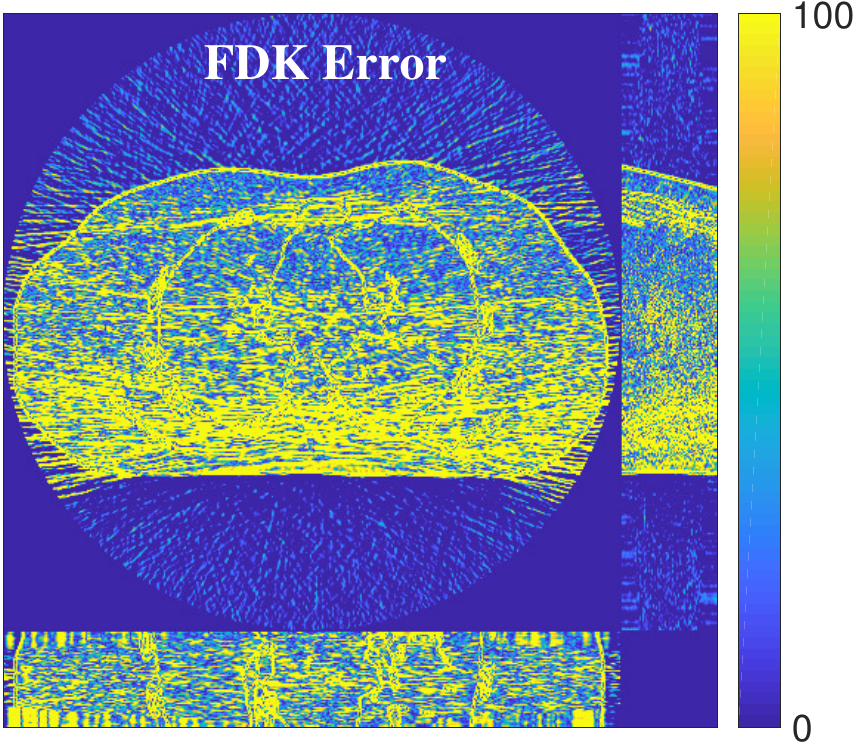}		
	\begin{tikzpicture}
	[spy using outlines={rectangle,green,magnification=2,size=10mm, connect spies}]
	\node {\includegraphics[width=0.21\textwidth]{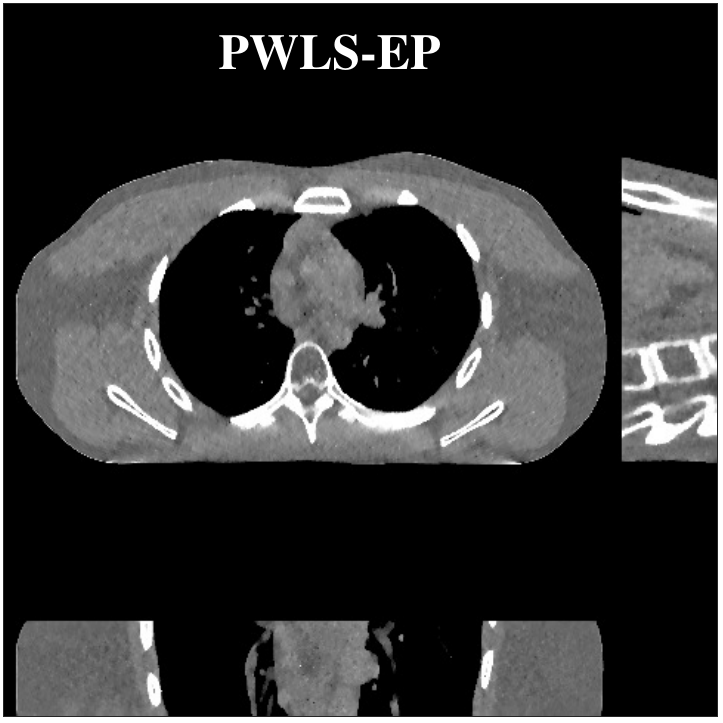}	};
	\spy on (0.87,0.12) in node [left] at (2.1,-1.2);		
	\spy on (-0.25,0.4) in node [left] at (2.1,1.65);		
	\end{tikzpicture}
	\includegraphics[width=0.25\textwidth]{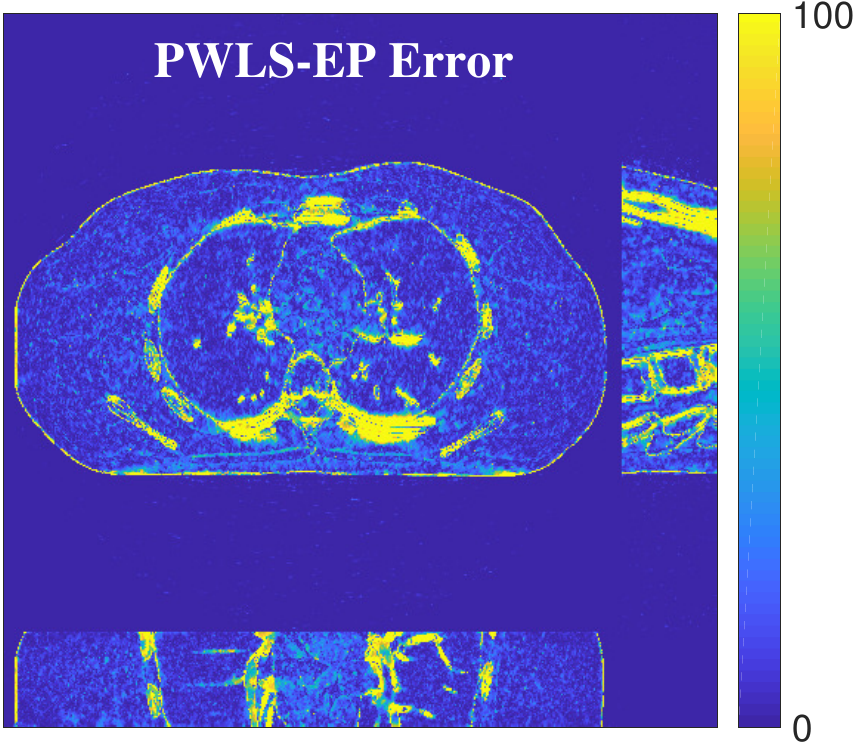}		
	
	\begin{tikzpicture}
	[spy using outlines={rectangle,green,magnification=2,size=10mm, connect spies}]
	\node {\includegraphics[width=0.21\textwidth]{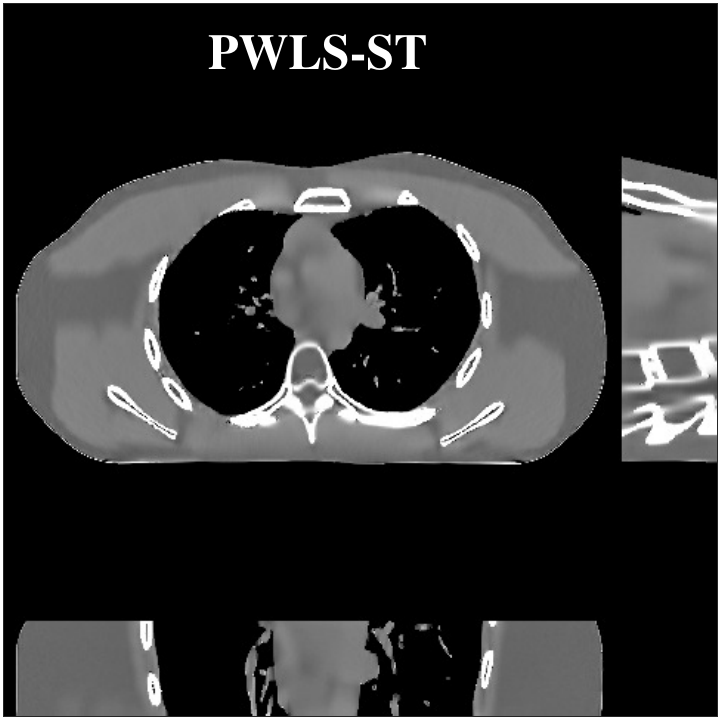}	};
	\spy on (0.87,0.12) in node [left] at (2.1,-1.2);	
	\spy on (-0.25,0.4) in node [left] at (2.1,1.65);			
	\end{tikzpicture}	
	\includegraphics[width=0.25\textwidth]{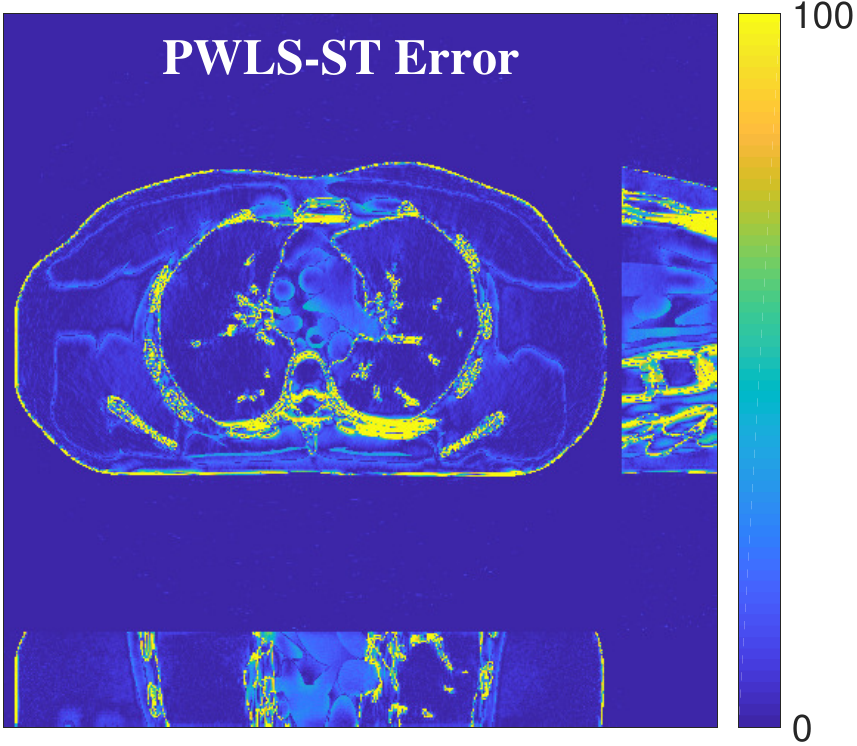}		
		
	\begin{tikzpicture}
	[spy using outlines={rectangle,green,magnification=2,size=10mm, connect spies}]
	\node {\includegraphics[width=0.21\textwidth]{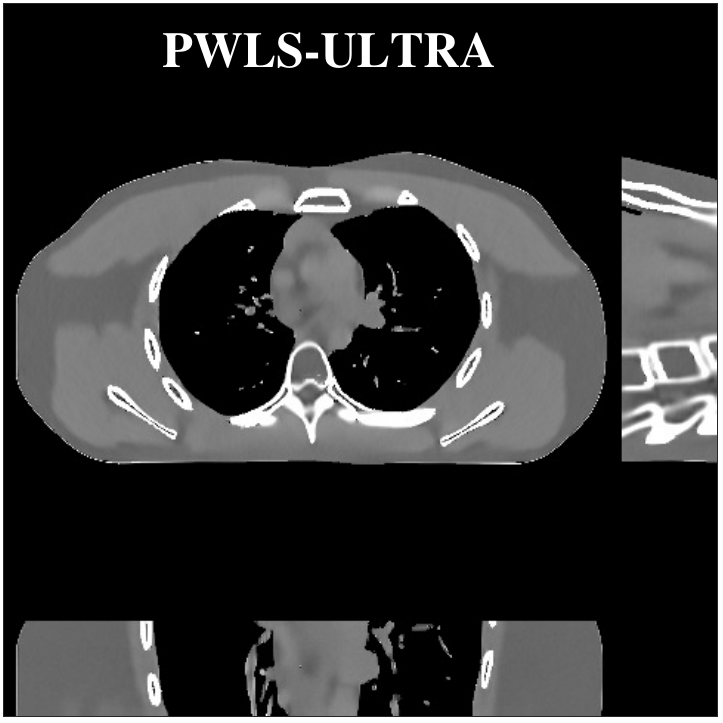}	};
	\spy on (0.87,0.12) in node [left] at (2.1,-1.2);		
	\spy on (-0.25,0.4) in node [left] at (2.1,1.65);		
	\end{tikzpicture}
	\includegraphics[width=0.25\textwidth]{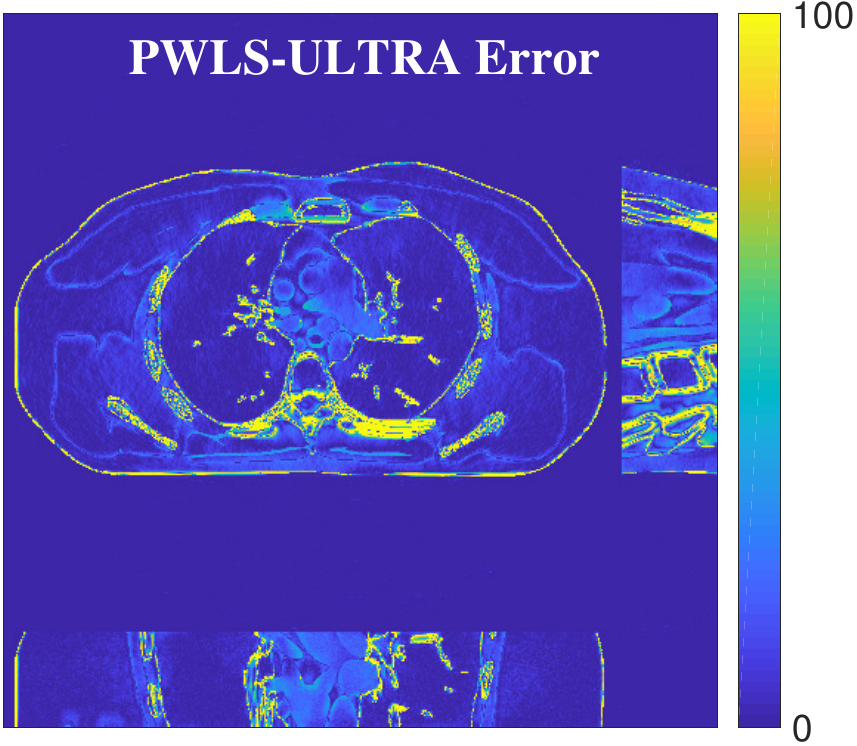}			
		\vspace{-0.2in}
	\caption{Comparison of the reconstructions and corresponding error images (shown for the central axial, sagittal, and coronal planes) for FDK, PWLS-EP, PWLS-ST ($\tau_j=1$, $\forall$ $j$), and PWLS-ULTRA ($K=15$) with patch-based weights $\tau_j$ at $I_0 = 5 \times 10^3$. The display window of reconstructions is $[800, 1200]$ HU. The unit of the display window of the error images is HU.}
	\label{fig:comp5e3}
	\vspace{-0.2in}
\end{figure}

%\begin{figure}[!t]
%	\setcounter{figure}{12}
%	\centering  	
%	\includegraphics[width=0.22\textwidth]{fig/5e3xfdk_new}		 
%	\includegraphics[width=0.255\textwidth]{fig/err5e3xfdk_new}		
%	\includegraphics[width=0.22\textwidth]{fig/5e3_l2b14dot5_os24_iter50}			
%	\includegraphics[width=0.255\textwidth]{fig/err5e3_l2b14dot5_os24_iter50}			
%	\includegraphics[width=0.22\textwidth]{fig/5e3block1_beta1dot5e5_gam20_SldDist2_iter2_os4_slice101_154_SldDist2_learn50}		    
%	\includegraphics[width=0.255\textwidth]{fig/err5e3block1_beta1dot5e5_gam20_SldDist2_iter2_os4_slice101_154_SldDist2_learn50}			
%	\includegraphics[width=0.22\textwidth]{fig/5e3kap1_block15_beta1dot2e4_gam20_SldDist2_clu20_iter2_os4_slice101_154_SldDist2_learn75}		    
%	\includegraphics[width=0.255\textwidth]{fig/err5e3kap1_block15_beta1dot2e4_gam20_SldDist2_clu20_iter2_os4_slice101_154_SldDist2_learn75}			
%	\caption{Comparison of the reconstructions and corresponding error images (shown for the central axial, sagittal, and coronal planes) for FDK, PWLS-EP, PWLS-ST ($\tau_j=1$, $\forall$ $j$), and PWLS-ULTRA ($K=15$) with patch-based weights $\tau_j$ at $I_0 = 5 \times 10^3$. }
%	\label{fig:comp5e3}
%		\vspace{-0.2in}
%\end{figure}

Section IV.E and Table III of \cite{zheng:17:pua} compared the performance of various methods for low-dose cone-beam (3D) CT reconstruction, for the XCAT phantom volume.
Fig.~\ref{fig:comp5e3} shows the reconstructions and the corresponding error (magnitudes) images (shown for the central axial, sagittal, and coronal planes)
at $I_0 = 5 \times 10^3$ for FDK, PWLS-EP, PWLS-ST (with $\tau_j=1$ $\forall$ $j$), and PWLS-ULTRA ($K=15$) with patch-based weights $\tau_j$.
PWLS-ULTRA provides better reconstructions and reconstruction errors compared to the conventional FDK and the non-adaptive PWLS-EP.
PWLS-ULTRA also outperforms the proposed PWLS-ST scheme, and provides sharper reconstructions of image edges (see zoom-ins).

\begin{figure*}[!t]
		\setcounter{figure}{13}
		\centering  	

\subfigure[]{\includegraphics[width=0.19\textwidth]{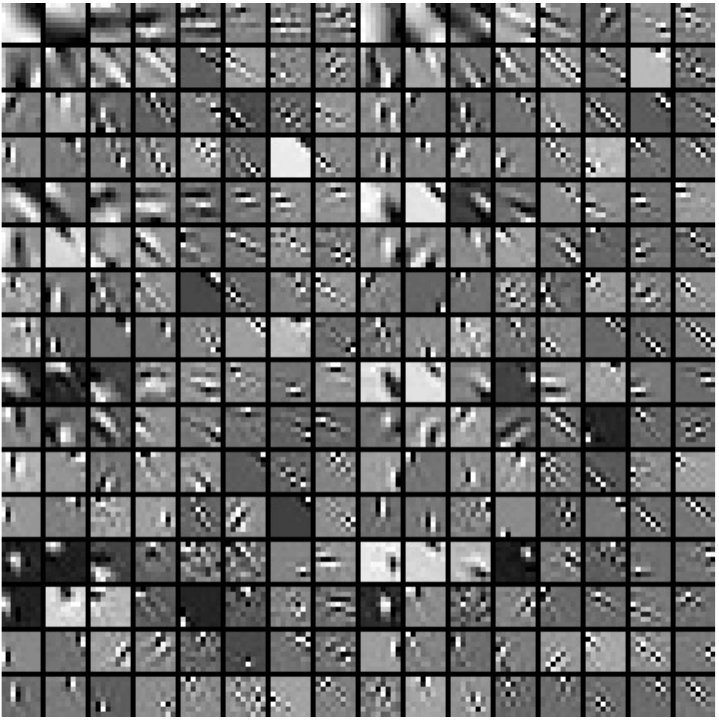}			
\includegraphics[width=0.19\textwidth]{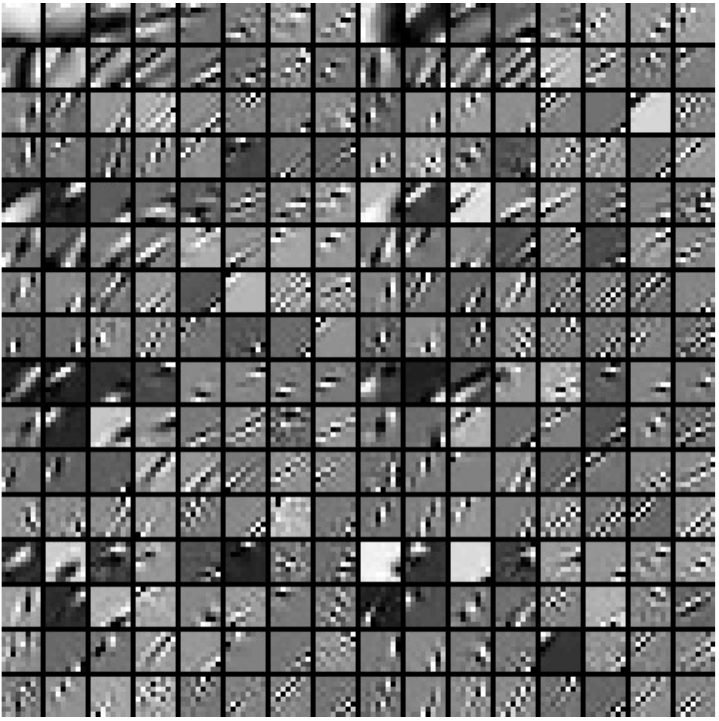}			
\includegraphics[width=0.19\textwidth]{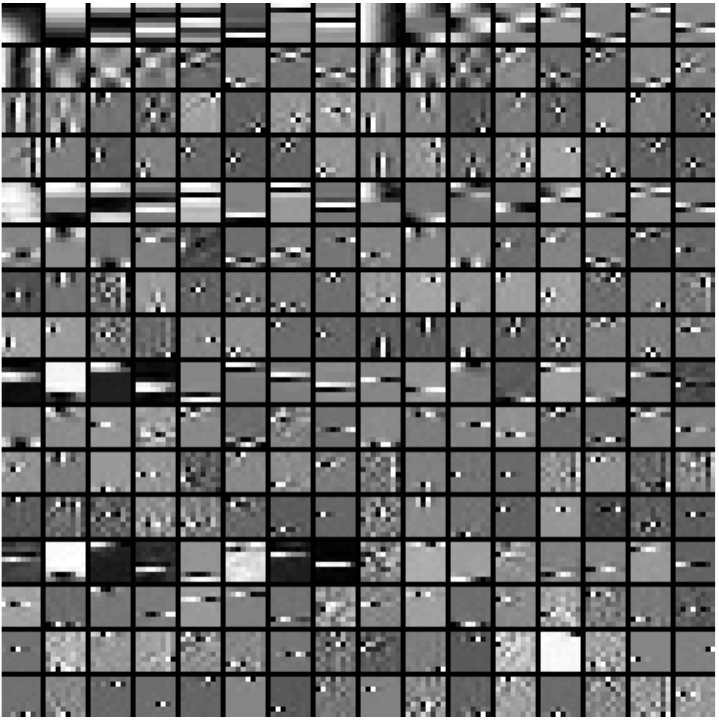}			
\includegraphics[width=0.19\textwidth]{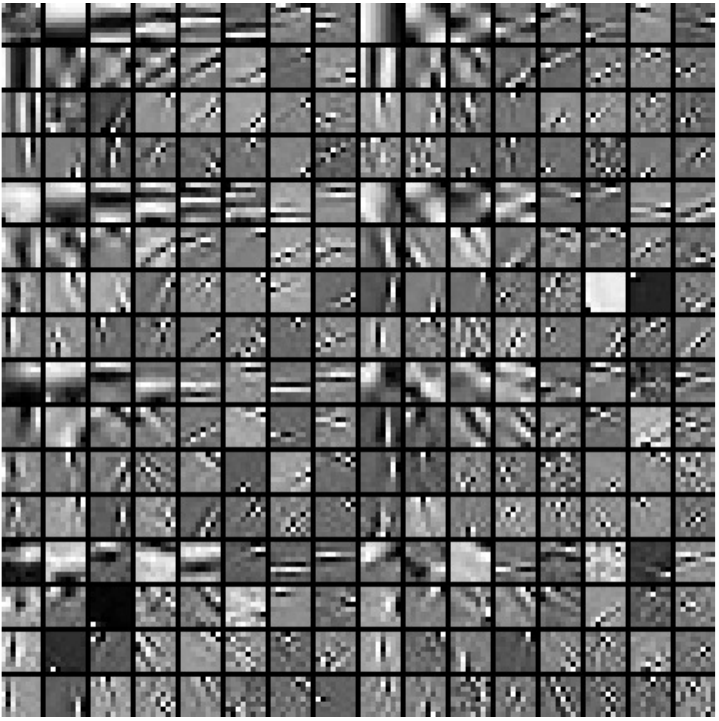}			
\includegraphics[width=0.19\textwidth]{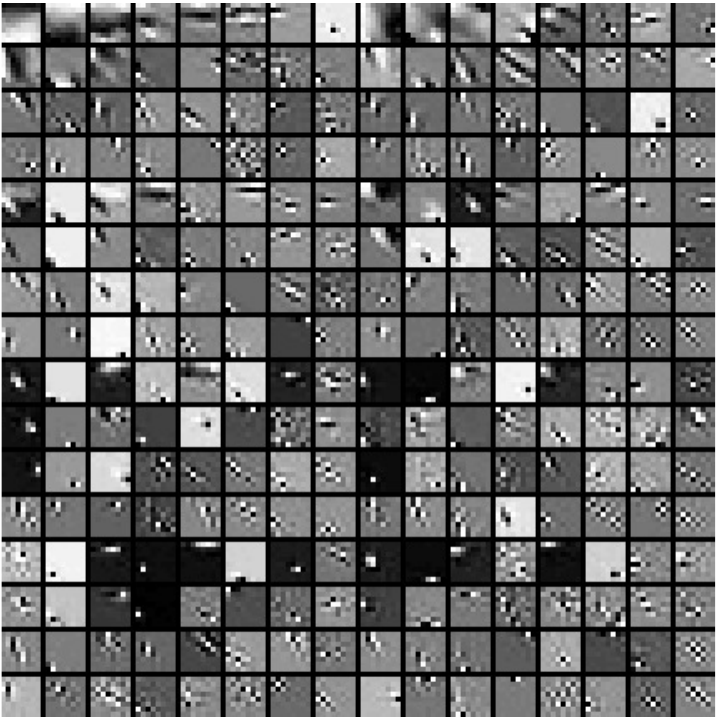}		}	

\subfigure[]{\includegraphics[width=0.19\textwidth]{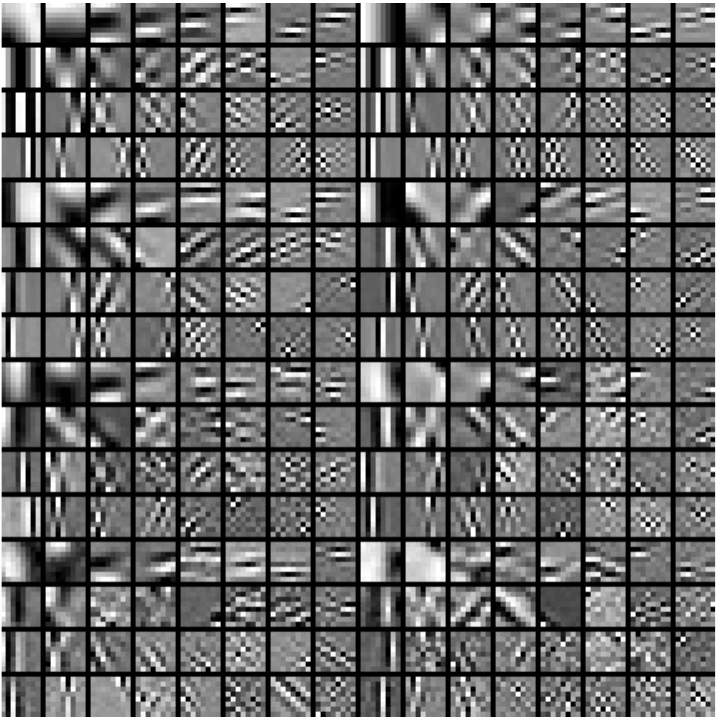}			
\includegraphics[width=0.19\textwidth]{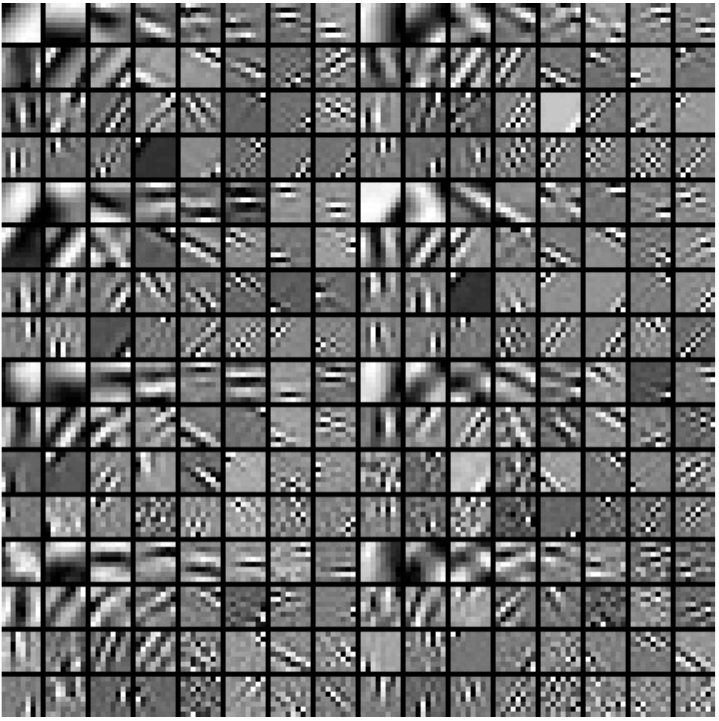}			
\includegraphics[width=0.19\textwidth]{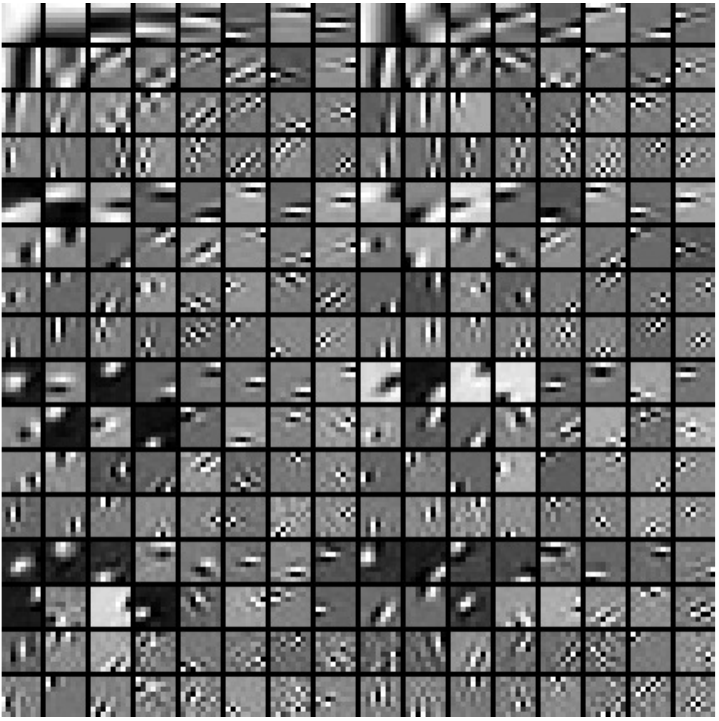}			
\includegraphics[width=0.19\textwidth]{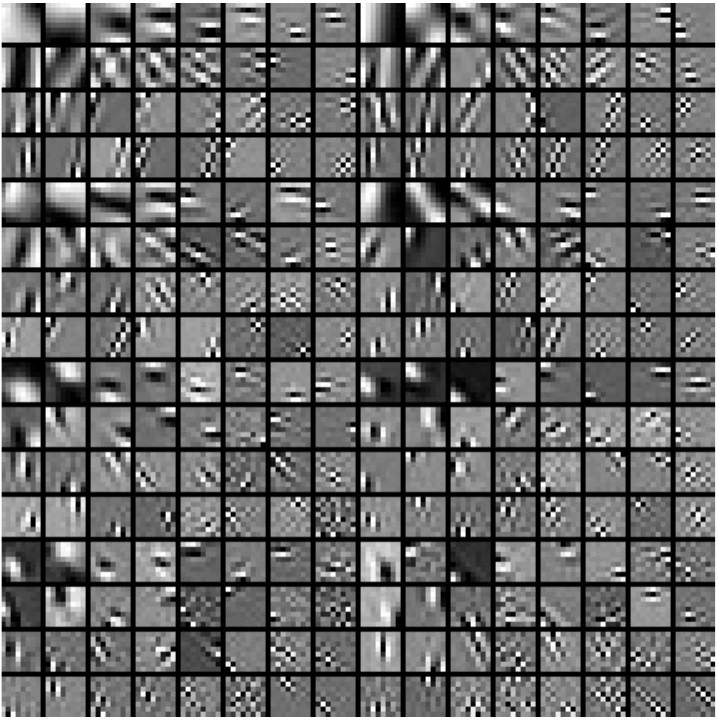}			
\includegraphics[width=0.19\textwidth]{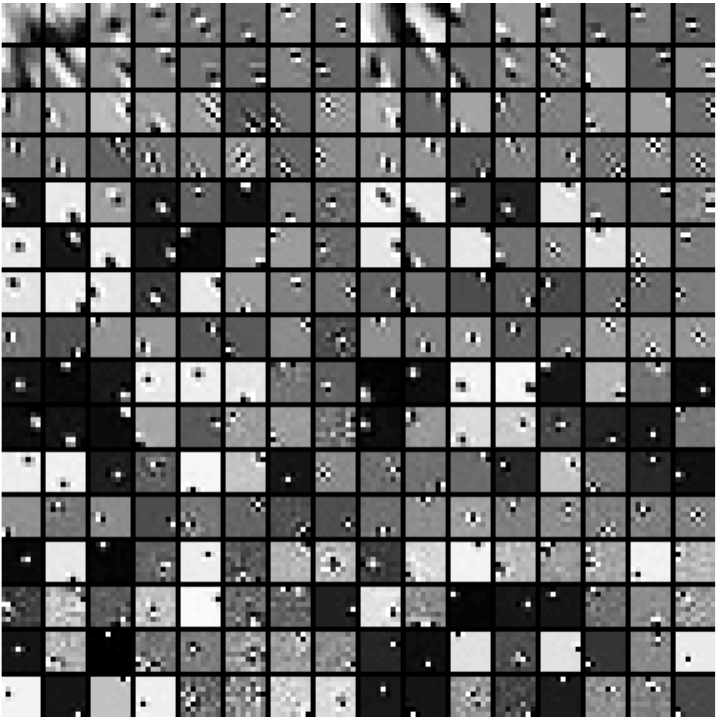}		}	

\subfigure[]{\includegraphics[width=0.48\textwidth]{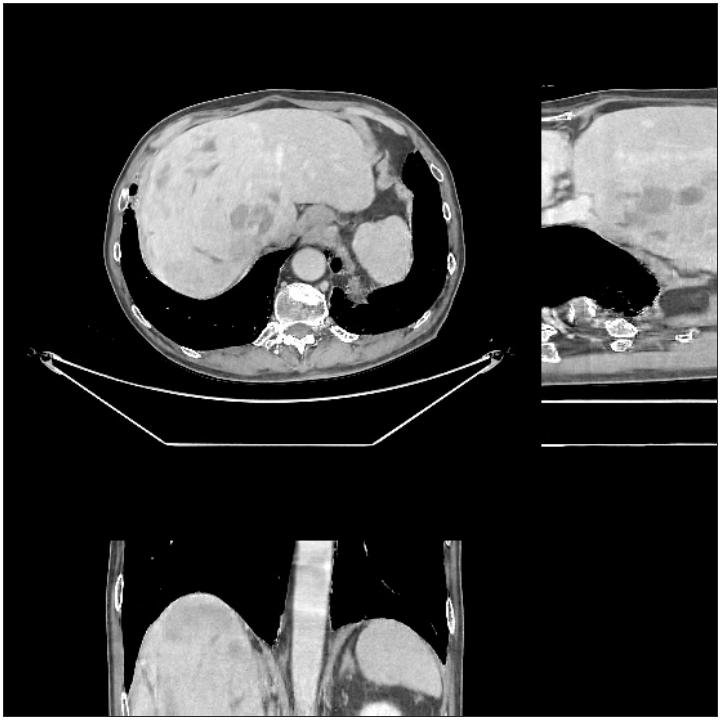}}
\subfigure[]{\includegraphics[width=0.48\textwidth]{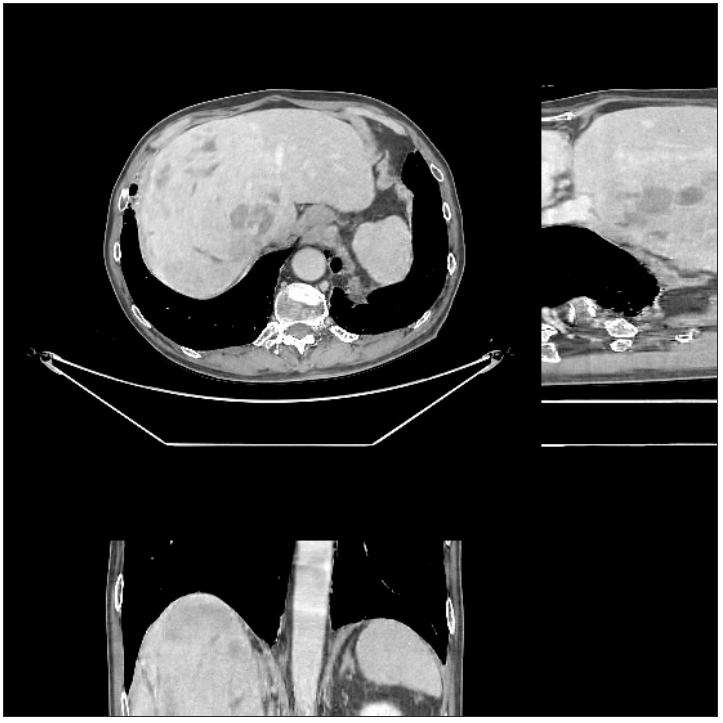} } 
		
	\caption{ The union of transforms learned (with $\eta=100$, $K = 5$) from (a) patches of the XCAT phantom and (b) from patches of the PWLS-EP reconstruction of the helical chest CT data are shown in the first and second rows, respectively. Only the first $8 \times 8$ slice of $256$ (among $512$) $8 \times 8 \times 8$ atoms are displayed. The corresponding PWLS-ULTRA-$\{\tau_j\}$ reconstructions (shown for the central axial, sagittal, and coronal planes) obtained with the transforms (a) and (b) are shown in (c) and (d), respectively. For the sagittal and coronal planes, we show the central $135$ out of $222$ axial slices.}
		\label{fig:chest_trans_comp}
	\end{figure*}

Recall that in Section IV.F, we used the transforms learned from the patches of the XCAT phantom volume to perform reconstruction of the chest volume from helical CT data.
Alternatively, one could learn the transforms from the patches of the PWLS-EP reconstruction of the helical CT data. Fig.~\ref{fig:chest_trans_comp} shows the union of transforms ($K=5$) learned from $8 \times 8 \times 8$ patches of the XCAT phantom and the PWLS-EP chest reconstruction, with $\eta=100$. These two union of transforms display some similar types of features, and provide similar visual reconstructions in PWLS-ULTRA (with patch-based weights $\tau_j$) in Fig.~\ref{fig:chest_trans_comp}.
Thus, the transform learning algorithm extracts quite general and effective sparsifying features for images, without requiring a very closely matched training dataset.

Fig.~\ref{fig:abdomen:ep_paras} provides abdomen reconstructions (shown for the central axial, sagittal, and coronal planes) from low-dose (120kVp, 150mA and 35mA) helical CT data for PWLS-EP with different regularization strengths. We have labeled the reconstruction with good trade-off between image resolution and noise in bold for both doses. These images were used to initialize the PWLS-ULTRA reconstructions in Section IV.F.

\begin{figure*}[!t]
			\setcounter{figure}{14}
	\centering  	
	\subfigure[$150$ mA, $\beta = 2^{17}$]{\includegraphics[width=0.32\textwidth]{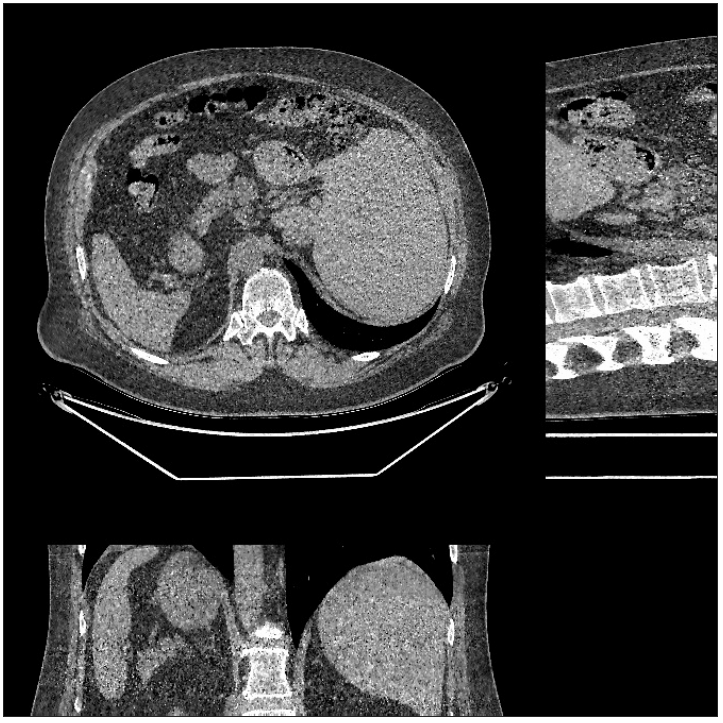}	}
	\subfigure[$\mathbf{150}$ \bf{mA}, $\boldsymbol{ \beta = 2^{18}}$]{\includegraphics[width=0.32\textwidth]{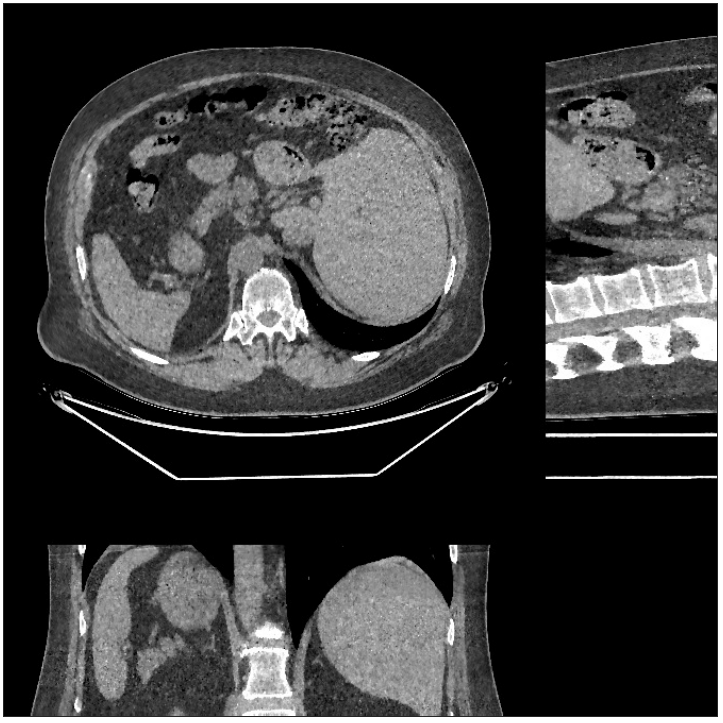}	}
	\subfigure[$150$ mA, $\beta = 2^{19}$]{\includegraphics[width=0.32\textwidth]{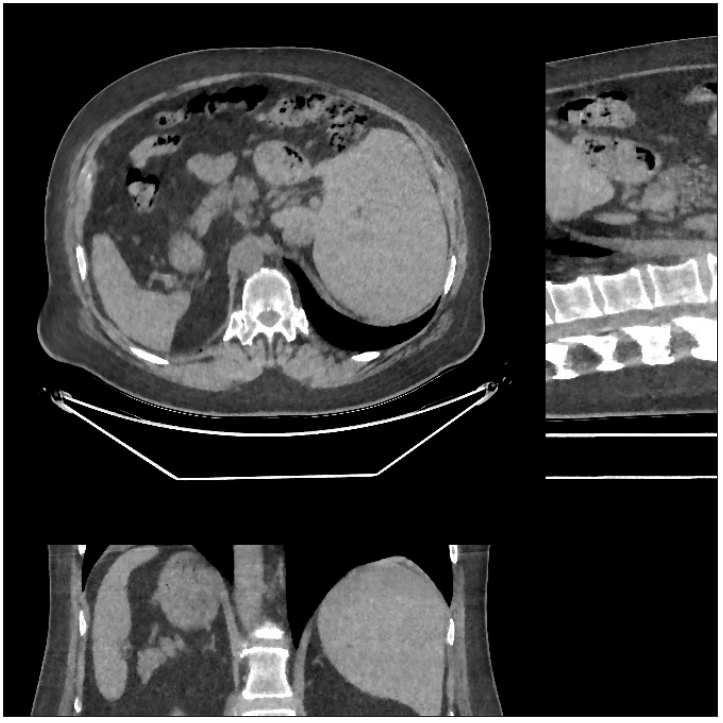}	}
	
	\subfigure[$35$ mA, $\beta = 2^{18}$]{\includegraphics[width=0.32\textwidth]{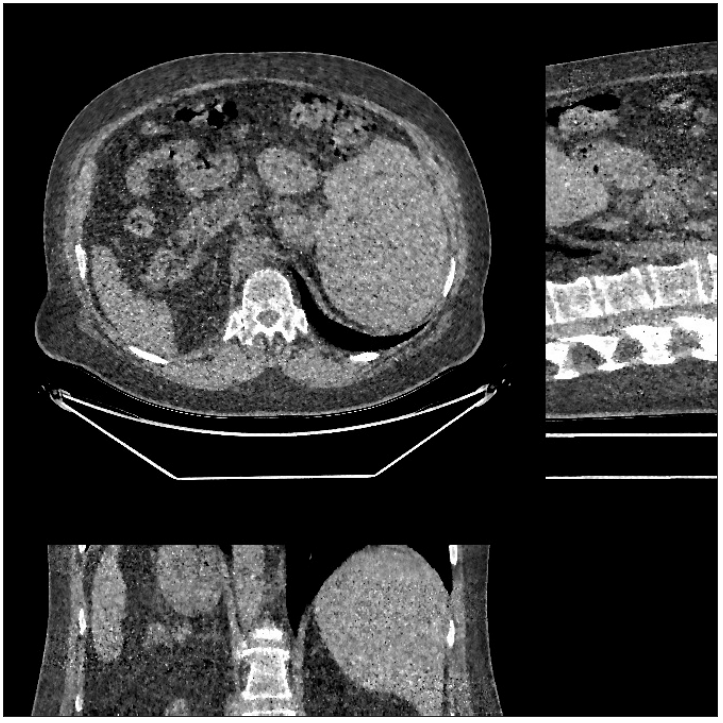}	}
	\subfigure[$35$ mA, $\beta = 2^{18.5}$]{\includegraphics[width=0.32\textwidth]{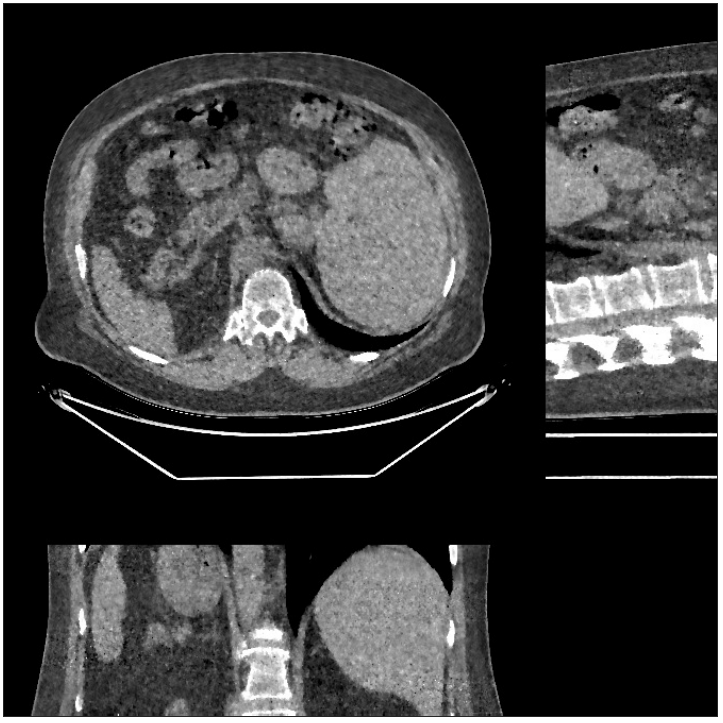}	}
	\subfigure[$\mathbf{35}$ \bf{mA}, $\boldsymbol{\beta = 2^{19}}$]{\includegraphics[width=0.32\textwidth]{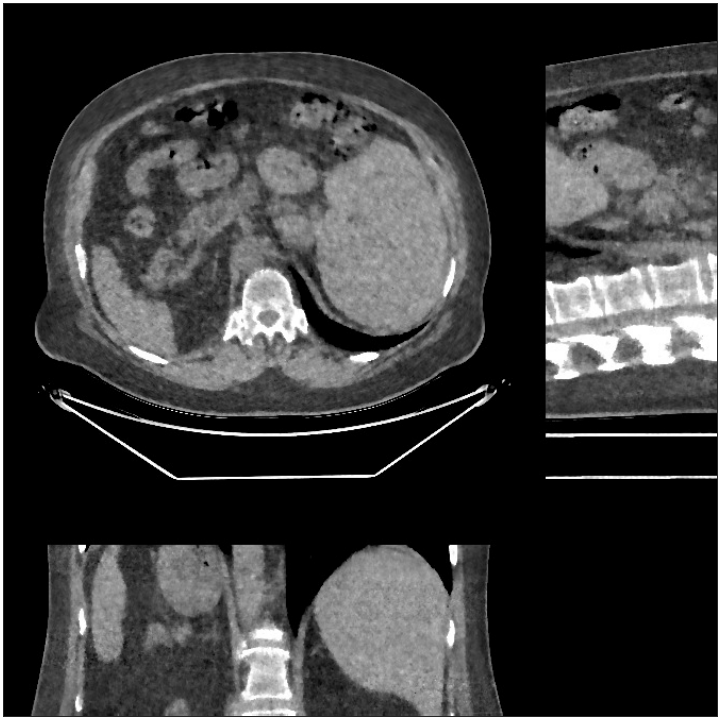}	}
%	\vspace{-0.15in}
	\caption{Abdomen reconstructions (shown for the central axial, sagittal, and coronal planes) from low-dose (120kVp, 150mA and 35mA) helical CT data for PWLS-EP with different regularization strengths.}
	\label{fig:abdomen:ep_paras}
%	\vspace{-0.15in}
\end{figure*}

\end{document}